%% file: icml.tex
\documentclass{article}  %

\usepackage{microtype}
\usepackage{graphicx}
\usepackage{subfigure}
\usepackage{booktabs} %

\usepackage{hyperref}

\usepackage[arxiv]{style/icml2024-latex/icml2024}

\usepackage{amsmath}
\usepackage{amssymb}
\usepackage{mathtools}
\usepackage{amsthm}

\input{commands}

\usepackage{multirow}
\usepackage{import}
\usepackage{listings}  %

\hypersetup{
    colorlinks,
    linkcolor={blue!50!black},
    citecolor={blue!50!black},
    urlcolor={blue!50!black}
}

\icmltitlerunning{GistScore: Learning Better Representations for In-Context Example Selection with Gist Bottlenecks}

\begin{document}

\twocolumn[

\icmltitle{\texorpdfstring{GistScore: Learning Better Representations for \\ In-Context Example Selection with Gist Bottlenecks}{GistScore: Learning Better Representations for In-Context Example Selection with Gist Bottlenecks}}

\icmlsetsymbol{atasapp}{$\dagger$}

\begin{icmlauthorlist}
\icmlauthor{Shivanshu Gupta}{xxx,atasapp}
\icmlauthor{Clemens Rosenbaum}{yyy}
\icmlauthor{Ethan R. Elenberg}{yyy}
\end{icmlauthorlist}

\icmlaffiliation{xxx}{University of California, Irvine, CA, USA}
\icmlaffiliation{yyy}{ASAPP, New York, NY, USA}

\icmlcorrespondingauthor{Shivanshu Gupta}{shivag5@uci.edu}

\icmlkeywords{Gisting, In-Context Learning, Large Language Models, Example Selection, Retrieval}

\vskip 0.3in
]

\printAffiliationsAndNotice{\textsuperscript{$\dagger$}Work done at ASAPP\quad}

\input{sections-icml/0_abstract}
\input{sections-icml/1_intro}

\input{sections-icml/7_related}

\input{sections-icml/2_setup}
\input{sections-icml/3_method}
\input{sections-icml/4_experiment}
\input{sections-icml/5_results}

\input{sections-icml/6_analysis}
\input{sections-icml/8_conclusion}
\clearpage
\input{sections-icml/9_required}

\bibliography{bibliography/anthology-pre2005,bibliography/anthology-post2005,bibliography/custom-rebiber}

\bibliographystyle{style/icml2024-latex/icml2024}

\clearpage

\appendix
\input{appendices-icml/set_extension}
\input{appendices-icml/datasets}
\input{appendices-icml/training}
\input{appendices-icml/all_results}

\end{document}

%% file: commands.tex
\newif\ifdebug

\newif\ifcomments
\commentstrue
\ifcomments
    \providecommand{\sg}[1]{{\protect\color{cyan}{[\textbf{shiv}: #1]}}}
    \providecommand{\ethan}[1]{{\protect\color{brown}{[\textbf{ethan}: #1]}}}
    \providecommand{\clemens}[1]{{\protect\color{red}{[\textbf{clemens}: #1]}}}
\else
    \providecommand{\sg}[1]{}
    \providecommand{\ethan}[1]{}
    \providecommand{\clemens}[1]{}
\fi

\newcommand{\tightparagraph}[1]{\smallbreak\noindent\textbf{#1}}

\definecolor{applegreen}{rgb}{0.01, 0.65, 0.01}
\definecolor{cardinal}{rgb}{0.77, 0.12, 0.23}
\definecolor{heldout}{rgb}{ .506,  .086,  .055}

\newcommand{\qnli}{QNLI}

\newcommand{\qnlickw}{wang-etal-2018-glue}
\newcommand{\mnli}{MNLI}

\newcommand{\mnlickw}{williams-etal-2018-broad}
\newcommand{\rte}{RTE}

\newcommand{\ssttwo}{SST2}

\newcommand{\ssttwockw}{socher-etal-2013-recursive}
\newcommand{\sstfive}{SST5}

\newcommand{\sstfiveckw}{socher-etal-2013-recursive}
\newcommand{\rotten}{Rotten Tomatoes}

\newcommand{\rottenckw}{pang-lee-2005-seeing}
\newcommand{\rteckw}{rte}
\newcommand{\mrpc}{MRPC}

\newcommand{\mrpcckw}{dolan-brockett-2005-automatically}
\newcommand{\qqp}{QQP}

\newcommand{\qqpckw}{wang-etal-2018-glue}
\newcommand{\paws}{PAWS}

\newcommand{\pawsckw}{zhang-etal-2019-paws}
\newcommand{\pawsx}{PAWSX}

\newcommand{\pawsxckw}{yang-etal-2019-paws}

\newcommand{\cmsqa}{CSQA}

\newcommand{\cmsqackw}{talmor-etal-2019-commonsenseqa}
\newcommand{\agnews}{AGNews}

\newcommand{\agnewsckw}{zhang2015character}
\newcommand{\smcalflow}{SMCalFlow}

\newcommand{\smcalflowckw}{andreas-etal-2020-task}
\newcommand{\smccs}{SMCalFlow-CS}

\newcommand{\mtop}{MTOP}

\newcommand{\mtopckw}{li-etal-2021-mtop}
\newcommand{\cogs}{COGS}

\newcommand{\cogsckw}{kim-linzen-2020-cogs}
\newcommand{\wanli}{WANLI}

\newcommand{\wanlickw}{liu-etal-2022-wanli}
\newcommand{\xnli}{XNLI}

\newcommand{\xnlickw}{conneau-etal-2018-xnli}
\newcommand{\mednli}{MedNLI}

\newcommand{\mednlickw}{herlihy-rudinger-2021-mednli}
\newcommand{\gsm}{GSM8K}

\newcommand{\gsmckw}{wei2023chainofthought}
\newcommand{\drop}{DROP}

\newcommand{\dropckw}{dua-etal-2019-drop}
\newcommand{\boolq}{BoolQ}

\newcommand{\boolqckw}{clark-etal-2019-boolq}
\newcommand{\cola}{CoLA}

\newcommand{\colackw}{warstadt-etal-2019-neural}
\newcommand{\tweet}{TweetEval}

\newcommand{\tweetckw}{barbieri-etal-2020-tweeteval}

\newcommand{\neo}{GPT-Neo-2.7B}
\newcommand{\neoemph}{\textbf{\neo}}

\newcommand{\llamaseven}{LLaMA-7B}
\newcommand{\llamasevenemph}{\textbf{\llamaseven}}
\newcommand{\llamathirteen}{LLaMA-13B}
\newcommand{\llamathirteenemph}{\textbf{\llamathirteen}}
\newcommand{\starcoder}{StarCoder}
\newcommand{\starcoderemph}{\textbf{\starcoder}}

\newcommand{\babbage}{Babbage}
\newcommand{\babbageemph}{\textbf{\babbage}}
\newcommand{\davinci}{Davinci}
\newcommand{\davinciemph}{\textbf{\davinci}}
\newcommand{\mistral}{Mistral}
\newcommand{\mistralemph}{\textbf{\mistral}}
\newcommand{\zephyr}{Zephyr}
\newcommand{\zephyremph}{\textbf{\zephyr}}

\newcommand{\rsc}{\textsc{Rand}}
\newcommand{\cossc}{\textsc{SBERT}}
\newcommand{\bmsc}{\textsc{BM25}}
\newcommand{\bsrsc}{\textsc{BSR}}

\newcommand{\setbsrsc}{\textsc{Set-BSR}}

\newcommand{\gistft}{\textsc{GS[F]}}
\newcommand{\gistmulti}{\textsc{GS[M]}}
\newcommand{\gistftok}[1]{\textsc{GS[F, {#1}]}}
\newcommand{\gistmtok}[1]{\textsc{GS[M, {#1}]}}
\newcommand{\setgistft}{\textsc{Set-}\gistft{}}
\newcommand{\setgistmulti}{\textsc{Set-}\gistmulti{}}

\newcommand{\epr}{EPR}
\newcommand{\epremph}{\textsc{\epr}}
\newcommand{\ceil}{CEIL}
\newcommand{\ceilemph}{\textsc{\ceil}}

\newcommand{\llmr}{LLM-R}
\newcommand{\llmremph}{\textsc{LLM-R}}

\newenvironment{resultstable}[1][]{
    \begingroup
    \setlength{\tabcolsep}{3pt} %
    \begin{table*}[#1]
    \centering
    \small
}{
    \end{table*}
    \endgroup
}

\newenvironment{resultstablesinglecol}[1][]{
    \begingroup
    \setlength{\tabcolsep}{3pt} %
    \begin{table}[#1]
    \centering
    \small
}{
    \end{table}
    \endgroup
}

%% file: sections-icml/0_abstract.tex
\begin{abstract}
In-context Learning (ICL) is the ability of Large Language Models (LLMs) to perform new tasks when conditioned on prompts comprising a few task examples. However, ICL performance can be critically sensitive to the choice of examples. To dynamically select the best examples for every test input, we propose \emph{Example Gisting}, a novel approach for training example encoders through supervised fine-tuning with an attention bottleneck between the inputs and outputs. These \emph{gist models} form the basis for \emph{GistScore}, a novel metric for scoring and selecting informative examples.
Further, we experiment with two variations: (1)~fine-tuning gist models for each dataset and (2)~multi-task training a single model on a large collection of datasets. The latter can be used for new tasks out-of-the-box, enabling a training-free ICL pipeline.
Evaluations with 21 datasets spanning 9 tasks and 8 diverse LLMs show that our fine-tuned models get state-of-the-art ICL performance with over 20\% absolute gain over off-the-shelf retrievers and 5\% over the best prior methods. Further, our multi-task model generalizes well to new tasks, datasets, and prompt templates. Selection using this model matches or outperforms prior methods while being three orders of magnitude faster than the strongest training-free baseline.\footnote{Code and models available at \url{https://github.com/Shivanshu-Gupta/gist-icl}.}

\end{abstract}

%% file: sections-icml/1_intro.tex
\section{Introduction}
\label{sec:intro}

\begin{figure}[ht!]
    \centering
    \includegraphics[width=\linewidth]{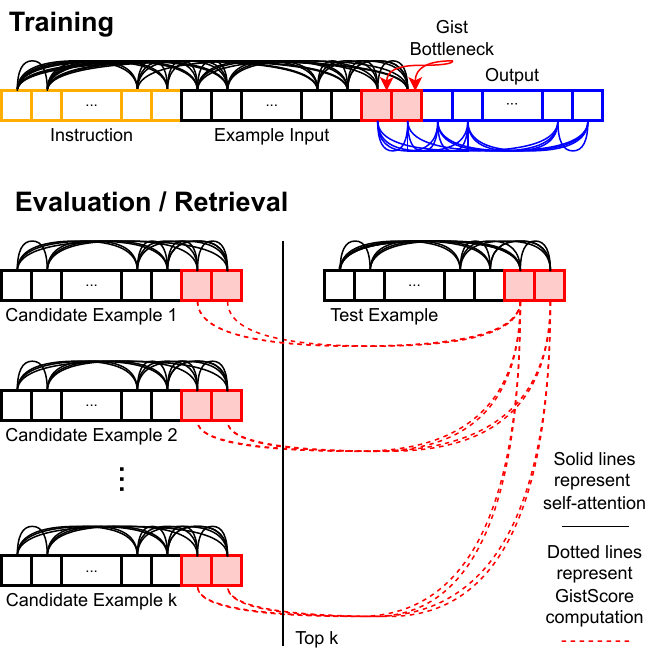}
    \caption{\textbf{Top} Example Gisting involves supervised training with an attention masking bottleneck. Here, gist tokens (red) may attend to example inputs (black) and the task instruction (yellow, optional), however, the output (blue) may only attend to the gist tokens. Training with such a bottleneck encourages concise, task-specifc encodings of salient aspects of inputs. Further, with multi-task training the model can be applied to new tasks out-of-the-box. \textbf{Bottom} Retrieval of the candidate examples with the highest GistScore with the test input (Task instruction omitted for brevity).
    }
    \label{fig:system_diagram}
\end{figure}

In-context Learning (ICL)~\citep{brown-etal-2020-fewshot} is a few-shot inference paradigm that leverages increasingly powerful large language models (LLMs) for new tasks by conditioning them on a prompt comprising a few task demonstrations. 
In contrast to traditional supervised fine-tuning, the training-free approach allows a single model to instantly switch between an arbitrary number of tasks with improved generaliztion~\citep{anil2022exploring,qiu-etal-2022-evaluating,drozdov2023compositional,wei2023chainofthought} and reasoning skills~\citep{wei2023chainofthought}.
Unfortunately, its performance is highly sensitive to the choice of examples placed in the prompt~\citep{zhao2021calibrate,liu-etal-2022-makes,lu-etal-2022-fantastically,rubin-etal-2022-learning,schick-schutze-2021-exploiting}.

Despite extensive prior work on better example selection methods \citep{rubin-etal-2022-learning,ye2023compositional,mualem2023submodular,gupta2023coveragebased}, the predominant approach in practice remains to use off-the-shelf retrievers like BM25 or cosine similarity between general-purpose encoder representations~\cite{reimers-2019-sentence-bert}. This is because the more effective prior approaches require training on the target task and/or training with feedback from a much larger Inference LLM~\cite{rubin-etal-2022-learning,ye2023compositional,hu2022incontext}, eliminating the key advantage of in-context learning.
More recently, \citet{gupta2023coveragebased} proposed training-free approaches based on BERTScore-Recall (BSR, \citet{zhang2020bertscore}).
However, BSR's quadratic complexity in input length makes it computationally expensive for long-text tasks. Further, its reliance on general-purpose encoders may lead to sub-optimal selection for many tasks.

To address these limitations, we seek to train computationally efficient retrievers that select informative in-context examples and can be used out-of-the-box for new tasks and datasets.
We propose \emph{Example Gisting}, a novel approach for training encoders for ICL example retrieval without feedback from a larger LLM. Based on Gisting, a recent technique by \citet{mu2023learning} for compressing prompts, Example Gisting induces an attention masking bottleneck between example inputs and outputs (Figure \ref{fig:system_diagram}, Top). Training with this bottleneck comprising a few \textit{gist} tokens forces the model to store task-specific salient input information into those tokens' activations. Subsequently, the trained gist model maps both candidate examples and new test inputs into sequences of gist token embeddings that can be used with \emph{GistScore}, a novel metric for scoring the informativeness of candidate examples (Figure \ref{fig:system_diagram}, Bottom).
By sharing BSR's functional form but operating on far fewer tokens, GistScore can be significantly faster while also being amenable to \citet{gupta2023coveragebased}'s extension to a set-level metric that can be used to find optimal sets of examples.
Finally, while fine-tuning on each dataset can yield optimal performance, to enable a training-free ICL pipeline we also experiment with multi-task training a single gist model on a large collection of datasets. By gisting instructions along with the example input, such a model can be used to select in-context examples for new tasks and datasets out-of-the-box.

Evaluating on 21 diverse datasets spanning 9 task categories and 8 diverse LLMs, we find that example selection using GistScore dramatically improves ICL. 
With fine-tuning, it consistently outperforms all prior selection methods, including ones that leverage task or LLM-specific training.
In particular, it beats off-the-shelf retrievers by up to 21 points and the best trained method by 5 points on average.
Further, our multi-task trained gist model recovers much of this performance gain.
Applied out-of-the-box, it can match prior trained methods and beat prior training-free methods even on held-out datasets and tasks while being thousands of times faster than BERTScore. 
Finally, congruent to \citet{gupta2023coveragebased}, we find that the set-extension of GistScore is highly effective for the task of Semantic Parsing and compositional generalization.
Our analysis shows that gist token embeddings capture abstract, task-specific salient aspects and can be effective for selection even for tasks that the gist model itself fails.
Overall, our multi-task trained gist model presents the best tradeoff of performance, ease of use, and selection speed and can potentially replace the standard approach of using off-the-shelf retrievers.

%% file: sections-icml/7_related.tex
\section{Related Work}
\label{sec:related}

\tightparagraph{In-Context Learning} Given the appeal of In-Context Learning as a training-free way to leverage LLMs, various example selection strategies to alleviate its sensitivity to choice of examples have been proposed: (1) selecting diverse examples to reduce redundancy among them~\citep{su2022selective,levy2022diverse,agrawal2022incontext,ye2022complementary}, (2) selecting examples that minimize the entropy of the LLM's output distribution for the test input~\citep{lu-etal-2022-fantastically,wu2023selfadaptive}, (3) Bayesian inference~\citep{wang2023large}, and (4) selecting examples as a set \citep{gupta2023coveragebased,ye2023compositional,mualem2023submodular}.

Perhaps the most relevant to our work are \citet{rubin-etal-2022-learning} and \citet{wang2023learning} which propose different ways to train example retrievers using feedback from a much larger LLM. However, task-specific finetuning requires abundant training data while also sacrificing the training-free nature of ICL limiting ease of use. Further, methods trained with an LLM-in-the-loop can lose effectiveness with larger Inference LLMs \citep{gupta2023coveragebased}. Orthogonally, \citet{gupta2023coveragebased} showed that simply using BERTScore-Recall (BSR) \citep{zhang2020bertscore} to score examples yields a training-free method that selects informative examples that demonstrate the \textit{salient aspects} of the test input. However, BSR requires matching every pair of token embeddings in the candidate and the test input making it computationally expensive for long-text tasks. Moreover, general-purpose encoders it uses may not capture informativeness for every task.

\tightparagraph{Attention and Memory} Example Gisting is inspired from \citet{mu2023learning}'s Gisting\footnote{We will refer to this method as Instruction Gisting to distinguish it from our proposed methods.} who use it to compress prompts. Both leverage attention bottlenecks to encode pertinent information in a few tokens, thereby acting as a memory. This is related to past work on improving memory and long-range sequence modeling with Transformers \citep{transformer_xl,sparse_transformer,longformer,rae2019compressive}. In particular, similar to the specialization of gist tokens, \citet{longt5} and \citet{streaming_transformer} model long sequence dependencies using specific tokens that act as a shared global memory, rather than passthrough tokens. Additionally, the sparsity induced by attention-masking in gisting is related to various sparse attention methods that have been proposed to improve Transformer efficiency. For example, \citet{transformer_xl} use block-wise dense local attention combined with recursive attention to the previous attention block. \citet{sparse_transformer} and \citet{longformer} use different forms of sliding (and strided) attention; they model long dependencies with either overlapping windows or specific tokens with overlapping attention.

%% file: sections-icml/2_setup.tex
\section{Preliminaries}
\label{sec:prelim}

\tightparagraph{In-context Learning (ICL)} LLMs have the ability to solve test inputs from new tasks when prompted with a few examples of that task. Formally, given a set of (input, output) pairs $\left\{\left(x_i, y_i\right)\right\}_{i=1}^k$, prompt template $\mathcal{T}$, and the test input $\mathbf{x}_{\text{test}}$, ICL using an \emph{Inference LLM} involves prompting it to conditionally generate the following test output:
\begin{equation}
    \mathbf{y}_{\text {test }} \sim \mathcal{P}_{LM}\left(\cdot \mid \mathcal{T}\left(\mathbf{x}_1, \mathbf{y}_1, \ldots, \mathbf{x}_k, \mathbf{y}_k, \mathbf{x}_{\text {test }}\right)\right)
\end{equation}

\tightparagraph{Example Selection}
This work focuses on the problem of selecting the $k$ in-context examples from a pool of $N \gg k$ labeled candidates. This is often necessary due to the limited context windows of LLMs. Moreover, even if it were possible to fit the entire pool in the prompt, LLMs have been shown to be highly sensitive to both the order~\citep{liu-etal-2022-makes} and the position of in-context examples~\citep{liu2023lost}. Thus, we seek to select the most relevant subset of candidates to improve both the computational efficiency and performance of ICL. Formally, the goal is to select a subset $\mathcal{S} \subset \left\{\left(x_i, y_i\right)\right\}_{i=1}^N$ of size $k$ that maximizes the probability of generating the desired $\mathbf{y}_{\text{test}}$ when the Inference LLM is conditioned on $\mathbf{x}_{\text {test}}$ and $\mathcal{S}$.

Beyond naïve random selection, the standard approach for this problem is to retrieve the top-$k$ examples from the candidate pool using either the BM25 algorithm \citep{robertson1993okapitrec,SprckJones2000APM} or dense retrieval using an off-the-shelf encoder. However, such general-purpose retrievers are not trained for selecting examples for in-context learning and can yield sub-optimal performance. Moreover, standard approaches for training retrievers \citep{karpukhin-etal-2020-dense} are not applicable as the gold retrieval is unknown.
As described in \S~\ref{sec:related}, prior approaches mitigate this by training with feedback from an Inference LLM \citep{rubin-etal-2022-learning,wang2023learning} or by using a more suitable general-purpose metric \citep{gupta2023coveragebased}.

\tightparagraph{Instruction Gisting} \citet{mu2023learning} proposed Instruction Gisting for compressing instruction-following prompts into shorter \emph{gists} for efficient LLM inference.
To perform this mapping, they train a gisting model, $GM$, to simultaneously compress prompts comprising task instructions into a few gist tokens and to follow instructions encoded in those gist tokens. This is achieved by masking attention such that any attention to/from the task instruction goes through the gist tokens.

Specifically, given an initial LM and an instruction tuning dataset $\mathcal{D}_{I} = \{(t_i, x_i, y_i)\}$ of instruction, (optional) input, and target tuples, the model is trained to predict $y$ from the sequence $\left[t, G, x\right]$, where $G$ is the sequence of special "gist" tokens added to the model vocabulary. Attention masking ensures that the model must predict based on the information of $t$ encoded in the activations above $G$. Denoting this gist of $t$ as $G(t)$, this approach of instruction tuning with a gist bottleneck can also be seen as distillation between a standard instruction-tuned LM and the gisting model $GM$:

\vspace{-5mm}\begin{align}
    \begin{split}
    & \mathcal{L}_G\left(p_G, \mathcal{D}_{I}\right) = \\
    & \operatornamewithlimits{\mathbb{E}}\limits_{t, x, y \sim \mathcal{D}_{I}}\left[\mathrm{KL}\left(p_{\mathrm{LM}}(y \mid t, x) \parallel  p_{GM}(y \mid G(t), x)\right)\right] .
    \end{split}
    \label{eq:instruction_gisting}
\end{align}\vspace{-3mm}

The trained gisting model can be used for new instructions by feeding it the sequence $\left[t, G\right]$, precomputing the activations above $G$, and then prompting it with those activations instead of $t$.

%% file: sections-icml/3_method.tex
\section{Method}
\label{sec:method}

\subsection{Intuition}
\label{sec:intuition}

The intuition behind our proposed approach relies on two insights from prior work.
First, \citet{streaming_transformer} showed that in attention-based architectures, arbitrary tokens can specialize to act as memory capturing essential information. \citet{mu2023learning} showed that this can be done in a targeted manner using bottlenecks induced via attention-masking.
Second, \citet{gupta2023coveragebased} showed that ICL example selection requires a relevance metric that can capture the potentially many task-specific facets under which two samples can be similar. These \emph{salient aspects} could be reasoning patterns, rules, or similar properties that make an example helpful for solving another.
These insights raise an interesting question: can we train attention-based models that can capture these salient aspects in memory-like bottlenecks? Specifically, we hypothesize that training models to perform tasks with a bottleneck between inputs and outputs would enable these bottlenecks to store the task-relevant aspects of the inputs, which are most helpful for distinguishing between candidates during example selection.

\subsection{Example Gisting}
\label{sec:example_gisting}

We now describe \emph{Example Gisting}, our approach to training example encoders for ICL example selection. Consider an initial LM and a labeled dataset for target task $t$: $\mathcal{D}_t = \{(x_i, y_i)\}$. Analogous to Instruction Gisting, we finetune a model $GM$ to predict $y_i$ given the inputs $[x_i, G]$, where $G$ is the attention bottleneck comprising $l$ gist tokens. As in Eq.~\eqref{eq:instruction_gisting}, this is akin to minimizing the following distillation objective:
\begin{align}
    \begin{split}
        & \mathcal{L}_G\left(p_G, \mathcal{D}_t\right) = \\
        &\operatornamewithlimits{\mathbb{E}}\limits_{x, y \sim \mathcal{D}_t}\left[\mathrm{KL}\left(p_{\mathrm{LM}}(y \mid x) \| p_{GM}(y \mid G(x))\right)\right] .
    \end{split}
\end{align}
As motivated in \S~\ref{sec:intuition}, Example Gisting trains the model to encode task-specific salient information of the inputs in the activations of the gist tokens. Next section will describe how gist activations can be used to select in-context examples. However, note that, unlike Instruction Gisting, example gists are only used to select examples for ICL, which can then be performed with any Inference LLM with the full text of the selected examples. Thus, Example Gisting is agnostic to the choice of Inference LLM and also does not suffer from the failure cases of Instruction Gisting, such as difficulty copying verbatim from the instruction \citep{mu2023learning}.

\subsection{Example Selection}
\label{sec:ex_selection}
A trained example gisting model can be used to select examples by mapping the candidates and the test input to sequences of \emph{gist embeddings} that can then be used to score the candidates. Specifically, given the gists $G(x_{\text{test}})$ of the test input and $G(z)$ for each candidate $z$, we use the final layer gist activations as gist embeddings, \emph{i.e.} $\mathbf{z} = \mathbf{z}_1, \ldots \mathbf{z}_l = G(z)[-1]$ and $\mathbf{x} = \mathbf{x}_1, \ldots \mathbf{x}_l = G(x_{\text{test}})[-1]$. Then we use the following metric, which we call \emph{GistScore}, to measure the relevance of each candidate with respect to the test input:
\begin{equation} \label{eq:gs}
    \mathrm{GS}(x, z) = \frac{1}{l}\sum\limits_{i=1}^{l} \max\limits_{j=1, \ldots l} \frac{\mathbf{x}_i^T \mathbf{z}_j}{\lVert\mathbf{x}_i\rVert \lVert\mathbf{z}_j\rVert}
\end{equation}
Finally, the top-$k$ examples with the highest GistScore are selected for ICL. Note that GistScore shares the functional form of BERTScore-Recall \citep{zhang2020bertscore}, and for $l = 1$, reduces to cosine similarity. $l > 1$ may be useful when a single embedding cannot encode all salient information. 
Further, as described in App~\ref{app:set}, GistScore admits \citet{gupta2023coveragebased}'s extension to a submodular set-level metric that can be greedily optimized to select examples together as a set.

\subsection{Multi-Task Training}
\label{sec:multitask_training}
While task-specific finetuning with the approach described in \S~\ref{sec:example_gisting} can yield greater performance, it shares the ease-of-use limitations prior to trained methods described in \S~\ref{sec:prelim}. To address we propose a multi-task training approach that enables the gisting model to be used out-of-the-box on new tasks without any additional training. This preserves the key advantage of ICL: the entire pipeline may be used with new tasks, domains, and prompt templates without any training. 

The key idea is to encode both the task instruction and the example input so that the model can distinguish the task and extract task-specific salient information from the input. Formally, given an initial LM and a collection of datasets $\mathcal{D}_{M} = \bigcup_{t \in T} \{(t, x, y): (x, y) \in \mathcal{D}_t\}$ spanning tasks $T$, we train the model to predict $y$ given the input sequence $[t, x, G]$ where $t$ is the task instruction and $G$ is the attention bottleneck as before. This is equivalent to minimizing the following multi-task distillation objective:
\begin{align}
\begin{split}
    & \mathcal{L}_G\left(p_G, \mathcal{D}_{M}\right)= \\
    &\operatornamewithlimits{\mathbb{E}}\limits_{t, x, y \sim \mathcal{D}_{M}}\left[\mathrm{KL}\left(p_{\mathrm{LM}}(y \mid t, x) \| p_{GM}(y \mid G(t, x))\right)\right] .
\end{split}
\end{align}\vspace{-8mm}

%% file: sections-icml/4_experiment.tex
\section{Experimental Setup}

\begingroup
\setlength{\tabcolsep}{2pt} %
\begin{table}[ht]
\centering
\small
\import{tables/tabulars/}{ds_summary_v2}
\caption{Datasets used in this work. \textcolor{heldout}{Red} highlights datasets held-out from our multi-task collection. We use the German and Russian splits of \xnli{}, Spanish and French of \pawsx{}, and IID and Compositional Generalization (CG) splits of \smcalflow{} and \cogs{}.
}
\label{tab:ds_summary}
\end{table}
\endgroup

\subsection{Datasets}
\label{sec:datasets}

\tightparagraph{Multi-task Corpus} For multi-task training gist models as described in \S~\ref{sec:multitask_training}, we use a subset of the FLAN 2022 collection \citep{longpre2023flan} which comprises 15M zero and few-shot prompts from over 473 datasets and 146 task categories.
Specifically, we subsample up to 10,000 zero-shot prompts at most 256 tokens long for every task category, yielding roughly 5M prompts.

\tightparagraph{ICL Evaluation} We evaluate on 21 datasets spanning 9 diverse task categories and multiple languages as listed in Table~\ref{tab:ds_summary}. These include several datasets not in FLAN-2022 to evaluate the out-of-the-box generalization of our multi-task gist models to new tasks, datasets, domains, etc.\footnote{Most of our held-in datasets also require the multi-task models to generalize to new prompt templates as our ICL prompt templates differ from FLAN-2022's.} In particular, \mednli{} \citep{\mednlickw} and \tweet{} \citep{\tweetckw} evaluate on held-out domains (Medical and Tweets) while \xnli{} \citep{\xnlickw} and \pawsx{} \citep{\pawsxckw} evaluate generalization to non-English languages.

We also evaluate on Semantic Parsing, a task that requires set-selection \citep{gupta2023coveragebased} and that is completely absent in our multi-task collection, making it a hard test of generalization for our multi-task models. Further, in addition to IID splits as for other datasets, for \smcalflow{} \citep{\smcalflowckw} and \cogs{} \citep{\cogsckw}, we also evaluate on compositional generalization (CG) splits. 
We include additional details about all the datasets, including splits, sample instances, selection and ICL templates, and metrics in App.~\ref{app:datasets}.

\subsection{Inference LLMs}
\label{sec:llms}
We experiment with eight diverse Inference LLMs including: 6 base LLMs viz. \neoemph{} \citep{black2021gptneo}, \llamasevenemph{} and \llamathirteenemph{} \citep{touvron2023llama}, \mistralemph{}\footnote{\url{https://hf.co/mistralai/Mistral-7B-v0.1}} \citep{jiang2023mistral}, OpenAI's \babbageemph{} (\texttt{babbage-002}) and \davinciemph{} (\texttt{davinci-002}); \zephyremph{}\footnote{\url{https://hf.co/HuggingFaceH4/zephyr-7b-alpha}} \citep{tunstall2023zephyr}, an instruction-tuned and aligned LLM; and \starcoderemph{}\footnote{\url{https://hf.co/bigcode/starcoder}} \citep{li2023starcoder}, a code-pretrained base LLM. \neo{}, \llamaseven{}, and \llamathirteen{} have context windows of 2048, \starcoder{} of 7000, \mistral{} and \zephyr{} of 8192, and \babbage{} and \davinci{} of 16384.

\subsection{Methods}
\label{sec:methods}

\subsubsection{GistScore}
\label{sec:gistdetails}

As described in \S~\ref{sec:method}, GistScore-based example selection involves training a gist model to produce example gists used within GistScore (\textsc{GS}) or its set-extension (\textsc{Set-GS}) for selection.
While decoder-only LMs can also be used for gisting \citep{mu2023learning}, we use encoder-decoder LMs. After example gisting training, we drop the decoder and retain only the encoder for gisting examples.
As described in \S~\ref{sec:example_gisting} and \S~\ref{sec:multitask_training}, we experiment with both finetuning gist models for each dataset as well as multi-task training a single model on the collection described in \S~\ref{sec:datasets} and then directly using it to gist and select in-context examples for downstream datasets.
We refer to these models as GistScore-based selection using these as \textsc{GS[Finetune]} or \textbf{\gistft{}} and \textsc{GS[Multitask]} or \textbf{\gistmulti{}}
and the set-extension as \textbf{\setgistft{}} and \textbf{\setgistmulti{}}, respectively.
For \gistft{}, we use \texttt{flan-t5-base} \citep{chung2022scaling} as the base LM and for \gistmulti{}, we use \texttt{flan-t5-large}. Each model is trained to produce gists of a fixed length $l$ denoted as \gistftok{$l$} and \gistmtok{$l$}.
We report results with $l=1$ unless specified otherwise.
Additional training details are provided in App.~\ref{app:training}.

\subsubsection{Baselines}
\label{sec:baselines}
In addition to randomly selecting in-context examples (\textbf{\rsc{}}), we compare with the following training-free ranking-based selection baselines: (1) dense retrieval using a general-purpose encoder (\texttt{all-mpnet-base-v2}) from SentenceBERT library (\textbf{\cossc{}}, \citet{reimers-2019-sentence-bert}), (2) sparse-retrieval using Okapi variant \citep{robertson1993okapitrec} of \textbf{\bmsc{}} from the \texttt{rank\_bm25}\footnote{\url{https://github.com/dorianbrown/rank_bm25}} library, and (3) BERTScore-Recall (\textbf{\bsrsc{}}, \citet{zhang2020bertscore}) using \texttt{deberta-large-mnli}\citep{williams-etal-2018-broad} as encoder. We also compare with the set-extension of \bsrsc{} (\textbf{\setbsrsc{}}) proposed by \citet{gupta2023coveragebased} for selecting optimal sets of examples.

Further, we compare with three methods that leverage training with feedback from an Inference LLM:
(1) \textbf{\epremph{}} \citep{rubin-etal-2022-learning} which uses LLM perplexity (\neo{}) to train a dense retriever for each dataset, (2) \textbf{\ceilemph{}} \citep{ye2023compositional} which uses \epremph{} and feedback from an LLM to train a Determinantal Point Process \citep{Kulesza_2012} for each dataset that is used to select examples as a set, and (3) \textbf{\llmremph{}} \citep{wang2023learning} which uses feedback from \llamaseven{} to train a reward model for evaluating candidate examples that is distilled into a dense retriever used for example selection.
For \epremph{} and \ceilemph{}, we compare with the 8-shot results reported in \citet{gupta2023coveragebased}, if available, and the 50-shot results from \citet{ye2023compositional}, otherwise. For \llmremph{}, we use their 8-shot ICL results with \llamaseven{}. Being multi-task trained, \llmr{} can also be applied to held-out tasks; however, as \citet{wang2023learning}'s held-out tasks are included in our multi-task collection, we only compare with it on its held-in datasets.

\subsection{Prompt Construction}
\label{sec:prompt}

Following prior work \citep{rubin-etal-2022-learning,gupta2023coveragebased}, for $k$-shot ($k=8$ unless specified otherwise) ICL with any given dataset, example selection method, and LLM, we construct the ICL prompt by selecting $k$ (or fewer depending on LLM context window) examples from the train split. (2) ordering the examples by increasing relevance so that the more relevant examples are closer to the test input, (3) linearizing the ordered examples and the test input using the dataset's ICL example template in Tables~\ref{tab:prompts_1}, \ref{tab:prompts_2}, and \ref{tab:prompts_3}, and (4) concatenating the linearizations.
For set-selection methods (\textsc{Set-BSR} and \textsc{Set-GS}), the examples are ordered by their corresponding instance-level score.

\begin{resultstablesinglecol}
    \import{tables/tabulars/}{main_singlecol}
    \caption{
    Average 8-shot ICL performance across all datasets with single-token GistScore and training-free baselines for different LLMs. See App. \ref{app:results} for complete results for each dataset and LLM. While finetuning (\gistft{}) yields the best performance, \gistmulti{} also outperforms the baselines and recovers much of \gistft{}'s performance despite requiring no finetuning.
    }
    \label{tab:main}
\end{resultstablesinglecol}

%% file: tables/tabulars/ds_summary_v2.tex
\begin{tabular}{@{}ll@{}}
\toprule
\textbf{Task Category} & \textbf{Dataset} \\ \midrule
\multirow{6}{*}{\shortstack[l]{Natural \\ Language \\ Inference}}
    & \qnli{} \cite{\qnlickw} \\
    & \mnli{} \cite{\mnlickw} \\
    & \rte{} \cite{\rteckw} \\
    & \textcolor{heldout}{\wanli{}} \cite{\wanlickw} \\
    & \textcolor{heldout}{\xnli{}} \cite{\xnlickw} \\
    & \textcolor{heldout}{\mednli{}} \cite{\mednlickw} \\ \midrule
\multirow{4}{*}{\shortstack[l]{Paraphrase \\ Detection}}
    & \mrpc{} \cite{\mrpcckw} \\
    & \qqp{} \cite{\qqpckw} \\
    & \paws{} \cite{\pawsckw} \\
    & \textcolor{heldout}{\pawsx{}} \cite{\pawsxckw} \\ \midrule
\multirow{2}{*}{\shortstack[l]{Question \\ Answering}}
    & \drop{} \cite{\dropckw} \\
    & \boolq{} \cite{\boolqckw} \\ \midrule
\multirow{3}{*}{\shortstack[l]{Semantic \\ Parsing}}
    & \textcolor{heldout}{\smcalflow{}} (SMC, \citet{\smcalflowckw} \\
    & \textcolor{heldout}{\mtop{}} \cite{\mtopckw} \\
    & \textcolor{heldout}{\cogs{}} \cite{\cogsckw} \\ \midrule
\multirow{3}{*}{\shortstack[l]{Sentiment \\ Analysis}}
    & \ssttwo{} \cite{\ssttwockw} \\
    & \sstfive{} \cite{\sstfiveckw} \\
    & \textcolor{heldout}{\rotten{}} \cite{\rottenckw} \\
    & \textcolor{heldout}{\tweet{}-emotion} \cite{\tweetckw} \\ \midrule
Commonsense
    & CommonSenseQA (\cmsqa{}, \citet{\cmsqackw}) \\ \midrule
CoT
    & \gsm{} \cite{\gsmckw} \\ \midrule
Summarization
    & \agnews{} \cite{\agnewsckw} \\ \midrule
\multirow{2}{*}{\shortstack[l]{Misc}}
    & \textcolor{heldout}{\tweet{}-offensive} \cite{\tweetckw} \\
    & \cola{} \cite{\colackw} \\\bottomrule
\end{tabular}

%% file: tables/tabulars/main_singlecol.tex
\begin{tabular}{lccccccc}
\toprule
\textbf{Selector} & \textbf{Neo} & \textbf{L7B} & \textbf{L13B} & \textbf{Mis.} & \textbf{Zeph.} & \textbf{Bab.} & \textbf{Dav.} \\
\midrule
\textbf{\rsc{}} & 38.0  & 46.3  & 48.9  & 56.4  & 58.8  & 39.9  & 52.4 \\
\textbf{\bmsc{}} & 46.2  & 53.6  & 57.3  & 64.0  & 65.1  & 45.4  & 57.4 \\
\textbf{\cossc{}} & 46.5  & 53.7  & 57.7  & 64.6  & 65.5  & 47.3  & 58.1 \\
\textbf{\bsrsc{}} & 57.1  & 60.8  & 64.6  & 70.9  & 70.1  & 57.3  & 65.4 \\
\midrule
\textbf{\gistmtok{1}} & 63.5  & 65.8  & 68.1  & 73.6  & 71.7  & 63.1  & 68.4 \\
\textbf{\gistftok{1}} & \textbf{68.1} & \textbf{70.1} & \textbf{71.8} & \textbf{76.5} & \textbf{74.9} & \textbf{67.3} & \textbf{71.0} \\
\bottomrule
\end{tabular}%

%% file: sections-icml/5_results.tex
\section{Results}
\label{sec:results}

\begin{figure}[t!]
    \centering
    \includegraphics[width=\linewidth]{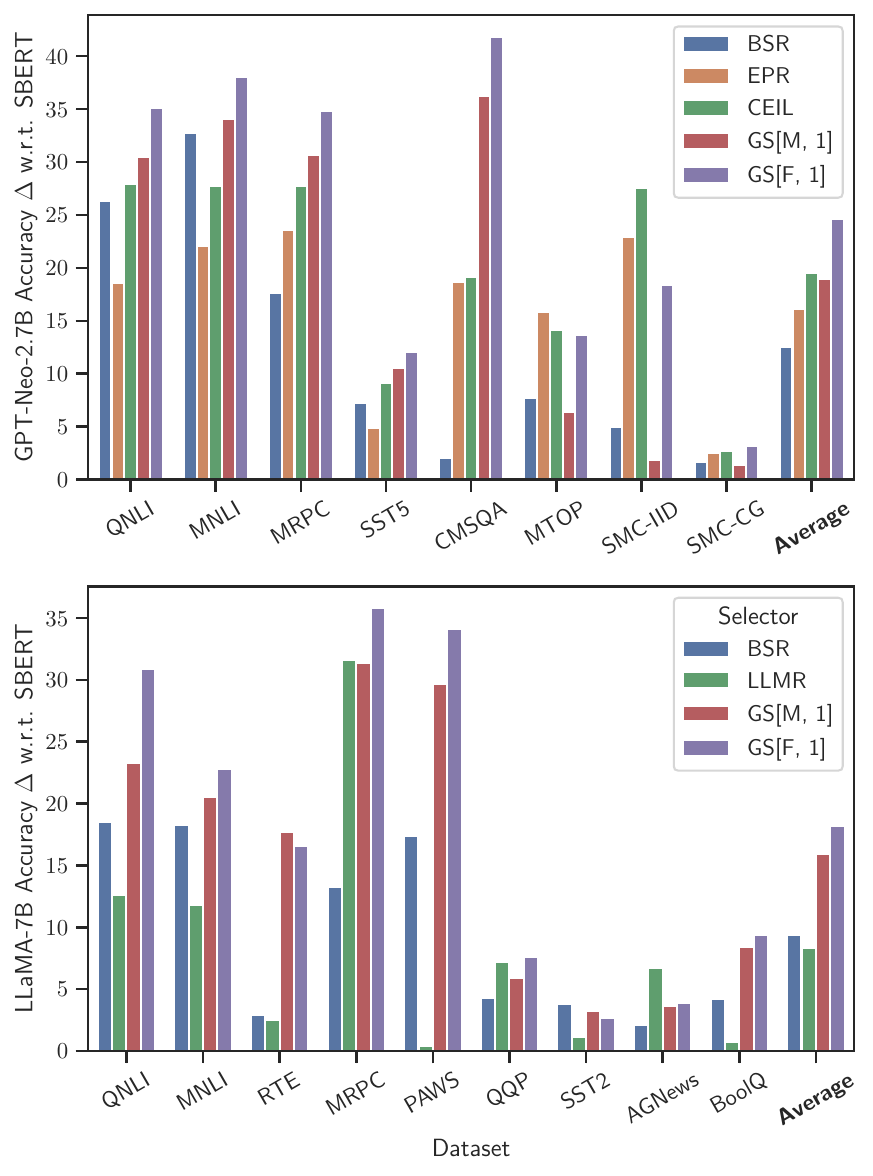}
    \caption{Single-token GistScore v/s \bsrsc{} and trained baselines: \epremph{} and \ceilemph{} with \neo{} \textbf{(Top)} and \llmremph{} with \llamaseven{} \textbf{(Bottom)}. All numbers are absolute gain in 8-shot ICL performance over \cossc{} except \epremph{} and \ceilemph{} on \mnli{}, \sstfive{}, \mrpc{}, and \cmsqa{} which are with 50 in-context examples.
    Both \gistft{} and \gistmulti{} consistently outperform all baselines, with \gistft{} performing the best. Semantic parsing is an exception as it requires additional gist tokens and set-selection (see Table \ref{tab:semparse}).}
    \label{fig:trained}
\end{figure}

\begin{figure}[t!]
    \centering
    \includegraphics[width=\linewidth]{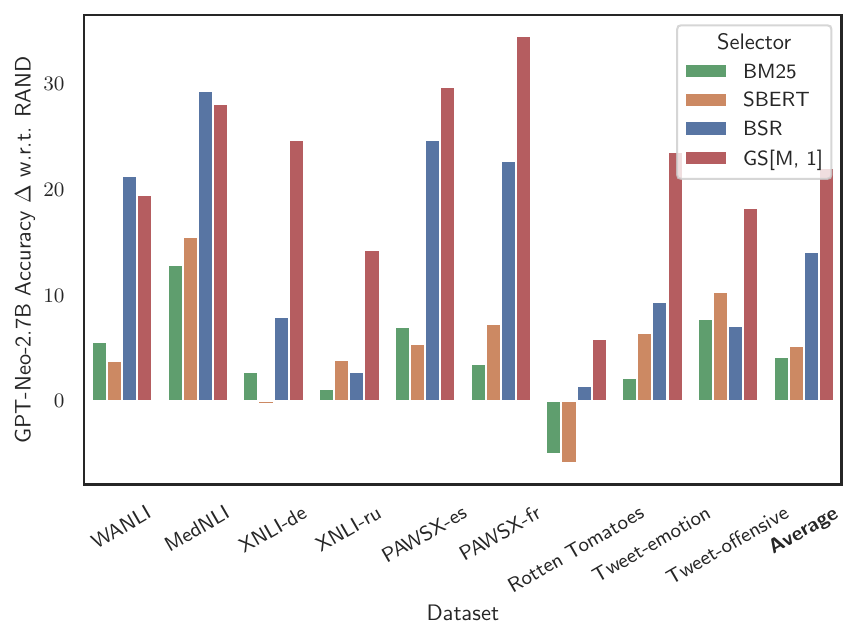}
    \caption{Comparison of training-free methods on held-out datasets. \gistmulti{} is able to generalize out-of-the-box to held-out datasets, domains (e.g., tweet, medical), and languages, significantly outperforming both off-the-shelf retrievers as well as the stronger but slower \bsrsc{}.}
    \label{fig:heldout}
\end{figure}

\begin{figure}[t!]
    \centering
    \includegraphics[width=\linewidth]{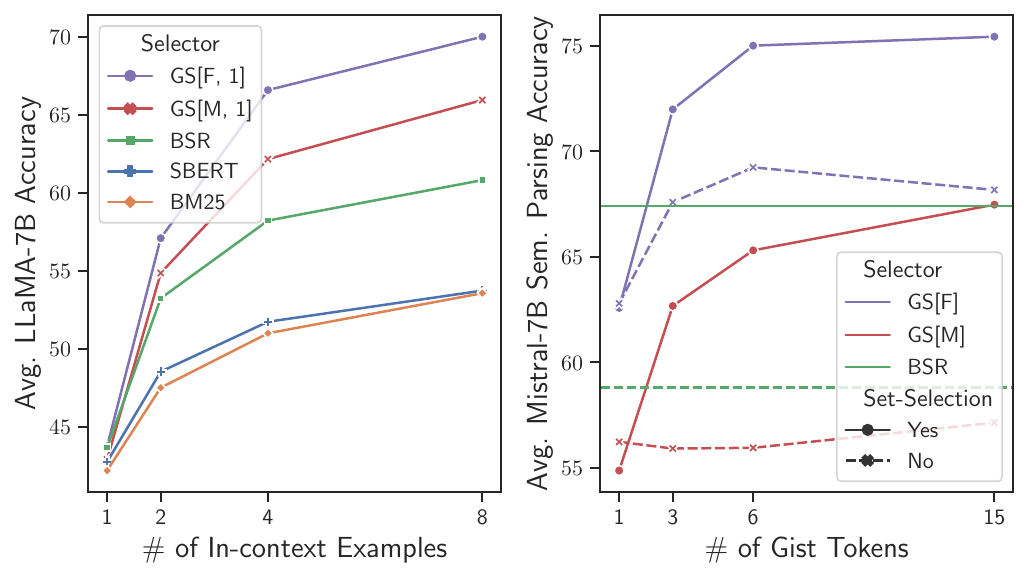}
    \caption{%
        \textbf{Left} \gistft{} and \gistmulti{} consistently outperform baselines across varying number of in-context examples, requiring just 2 examples to surpass 8-shot ICL using \cossc{} and \bmsc{}.
        \textbf{Right} Due to their complex compositional nature, Semantic Parsing datasets benefit from additional gist tokens and set-selection. With 15 tokens, \setgistmulti{} matches the average 8-shot semantic parsing ICL performance of \setbsrsc{}, while \setgistft{} vastly outperforms it. See Table \ref{tab:semparse} for trained baselines and Table \ref{tab:set-all} for complete results.
    }
    \label{fig:shots_set}
\end{figure}

\begin{resultstablesinglecol}
    \import{tables/tabulars/heldout}{semparse-starcoder}
    \caption{8-shot ICL using \starcoder{} for Semantic Parsing datasets with independent ranking (top) and set selection (bottom) methods. 15-token \setgistft{} outperforms all baselines while \setgistmulti{} matches them despite never being trained for Semantic Parsing.}
    \label{tab:semparse}
\end{resultstablesinglecol}

\tightparagraph{Finetuned GistScore is the superior method in-context example-selection method.}
Table \ref{tab:main} and Figure \ref{fig:trained} compare the performance of ICL example selection using single-token GistScore with prior training-free and trained approaches for a variety of datasets and Inference LLMs.
Additional results for all datasets and LLMs are provided in App.~\ref{app:results}.
With the exception of Semantic Parsing datasets, \gistftok{1} consistently and dramatically outperforms all baselines, beating the training-free \cossc{} and \bsrsc{} by up to 21 and 11 points and the trained baselines, \ceilemph{} and \llmremph{}, by 5 and 8 points on average, respectively.
Finally, the gains from GistScore persist across varying number of in-context examples (Figure \ref{fig:shots_set}, Left) -- with just 2 examples, it outperforms 8-shots retrieved using general-purpose retrievers.

\tightparagraph{Semantic Parsing benefits from additional gist-tokens and set-selection.}
While a single gist token works best for most datasets, it can be insufficient to capture all the salient information in complex compositional semantic parsing instances. Moreover, as shown in \citep{gupta2023coveragebased}, their compositional nature also necessitates set-selection as opposed to independent ranking-based selection, which can yield redundant examples while omitting information.
Indeed, as shown in Figure \ref{fig:shots_set} (Right), set-selection of examples using the set-extension of GistScore with additional gist-tokens leads to dramatic gains for these datasets for both variants of gist models. In fact, with 15 tokens, \setgistft{} outperforms all prior methods on semantic parsing as well (see Table \ref{tab:semparse}).

\tightparagraph{Multi-task training yields strong performance out-of-the-box.}
Table \ref{tab:main} and Figure \ref{fig:trained} show that without task-specific finetuning, example selection using our multi-task trained gist model (\gistmulti{}) is able to recover much of the performance of its finetuned counterparts (\gistft{}) and matches or outperforms all baselines including trained ones like \epremph{}, \ceilemph{}, and \llmremph{}.
Further analyzing its held-out performance in Figure \ref{fig:heldout}, we see that \gistmulti{} is able to generalize out-of-the-box to held-out datasets, domains (e.g., tweet, medical), and languages, significantly outperforming off-the-shelf retrievers (\bmsc{} and \cossc{}) as well as the stronger \bsrsc{}.
For semantic parsing datasets, unlike \gistft{}, just using more gist tokens without set-selection was only marginally effective (Figure \ref{fig:shots_set}, Right), likely because semantic parsing is not included in the multitask corpus, and so the gist model is unable to leverage the additional tokens well. However, when used for set-selection, additional tokens dramatically improve performance --- with 15 tokens, despite not being trained for semantic parsing, \setgistmulti{} is able to match even trained methods like \epremph{} and \ceilemph{} (see Table \ref{tab:semparse}).
These results confirm that multi-task training for gisting both task instructions and example inputs enables generalization to new tasks making it a promising approach for an improved training-free ICL pipeline.

\begin{figure}
    \centering
    \includegraphics[width=\linewidth]{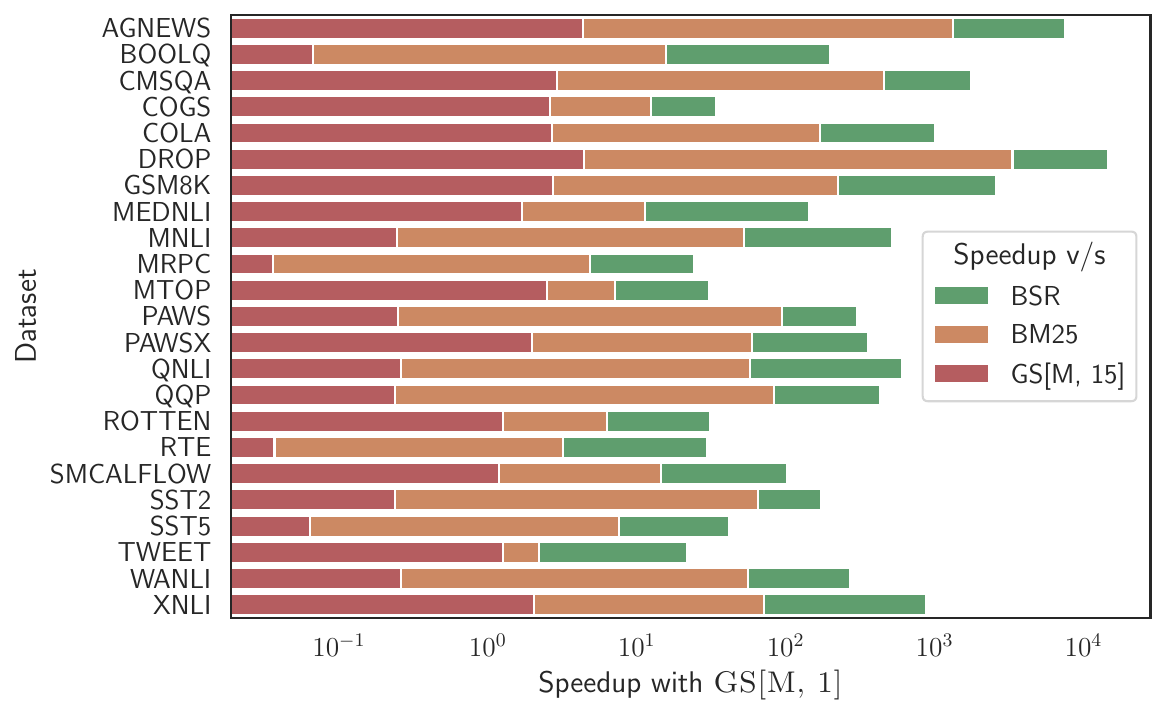}
    \caption{Example selection using GistScore (\gistmtok{1}) is up to four (three) orders of magnitude faster than \bsrsc{} (\bmsc{}), and scales well with the number of gist tokens.}
    \label{fig:speedup}
\end{figure}

\tightparagraph{Selection using GistScore is significantly faster than \bsrsc{}.}
Despite sharing its functional form and hence quadratic time-complexity in number of tokens, GistScore can be faster than \bsrsc{} as it compares only a few gist tokens. Figure \ref{fig:speedup} shows that this yields thousands of times faster selection with single-token GistScore compared to \bsrsc{}, which took over 20 seconds per test input for some datasets (see Table \ref{tab:speeds}). Further, due to GPU acceleration, we found GistScore to be significantly faster than even \bmsc{}.

%% file: tables/tabulars/heldout/semparse-starcoder.tex
\begin{tabular}{lcccccc}
\toprule
\multirow{2}[1]{*}{\textbf{Selector}} & \multicolumn{2}{c}{\textbf{SMC}} & \multicolumn{2}{c}{\textbf{\cogs{}}} & \multirow{2}[1]{*}{\textbf{\mtop{}}} & \multirow{2}[1]{*}{\textbf{AVG}} \\
     & \textbf{IID} & \textbf{CG} & \textbf{IID} & \textbf{CG} &      &  \\
\midrule
\textbf{\bsrsc{}} & 65.3 & 18.6 & 91.8 & 78.0   & 68.0   & 64.3 \\
\textbf{\epremph{}} & 69.8 & 17.3 &      &      & 72.6 &  \\
\textbf{\textsc{GS[M, 1]}} & 58.2 & 16.0   & 88.4 & 70.8 & 68.5 & 60.4 \\
\textbf{\textsc{GS[F, 1]}} & 69.0   & 14.6 & 89.0   & 75.0   & 71.0   & 63.7 \\
\midrule
\textbf{\textsc{Set-BSR}} & 69.6 & 51.4 & 92.4 & 77.1 & 70.0   & 72.1 \\
\textbf{\ceilemph{}} & 71.0   & 31.8 &      &      & 73.7 &  \\
\textbf{\textsc{Set-GS[M, 15]}} & 69.2 & 52.3 & 91.7 & 71.6 & 71.7 & 71.3 \\
\textbf{\textsc{Set-GS[F, 15]}} & \textbf{73.7} & \textbf{53.1} & \textbf{94.7} & \textbf{81.4} & \textbf{75.5} & \textbf{75.7} \\
\bottomrule
\end{tabular}%

%% file: sections-icml/6_analysis.tex
\section{Analysis}
\label{sec:analysis}

\begin{figure}
    \centering
    \includegraphics[width=0.8\linewidth]{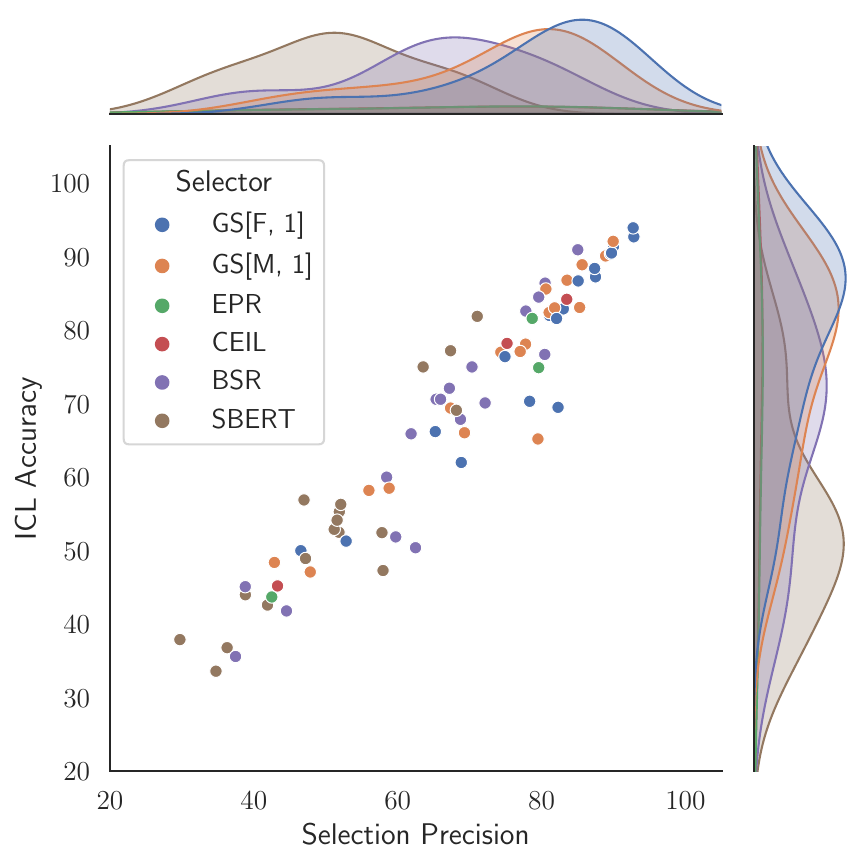}
    \caption{ICL accuracy (using \neo{} here) across all classification tasks is strongly correlated with the precision of the various selectors, \emph{i.e.} per-dataset-average of the fraction of in-context examples with the same label as the test input. This suggests that retrieving such examples is the primary driver of ICL performance for these datasets.
    }
    \label{fig:precision_corr}
\end{figure}

\begin{figure}
    \centering
    \includegraphics[width=\linewidth]{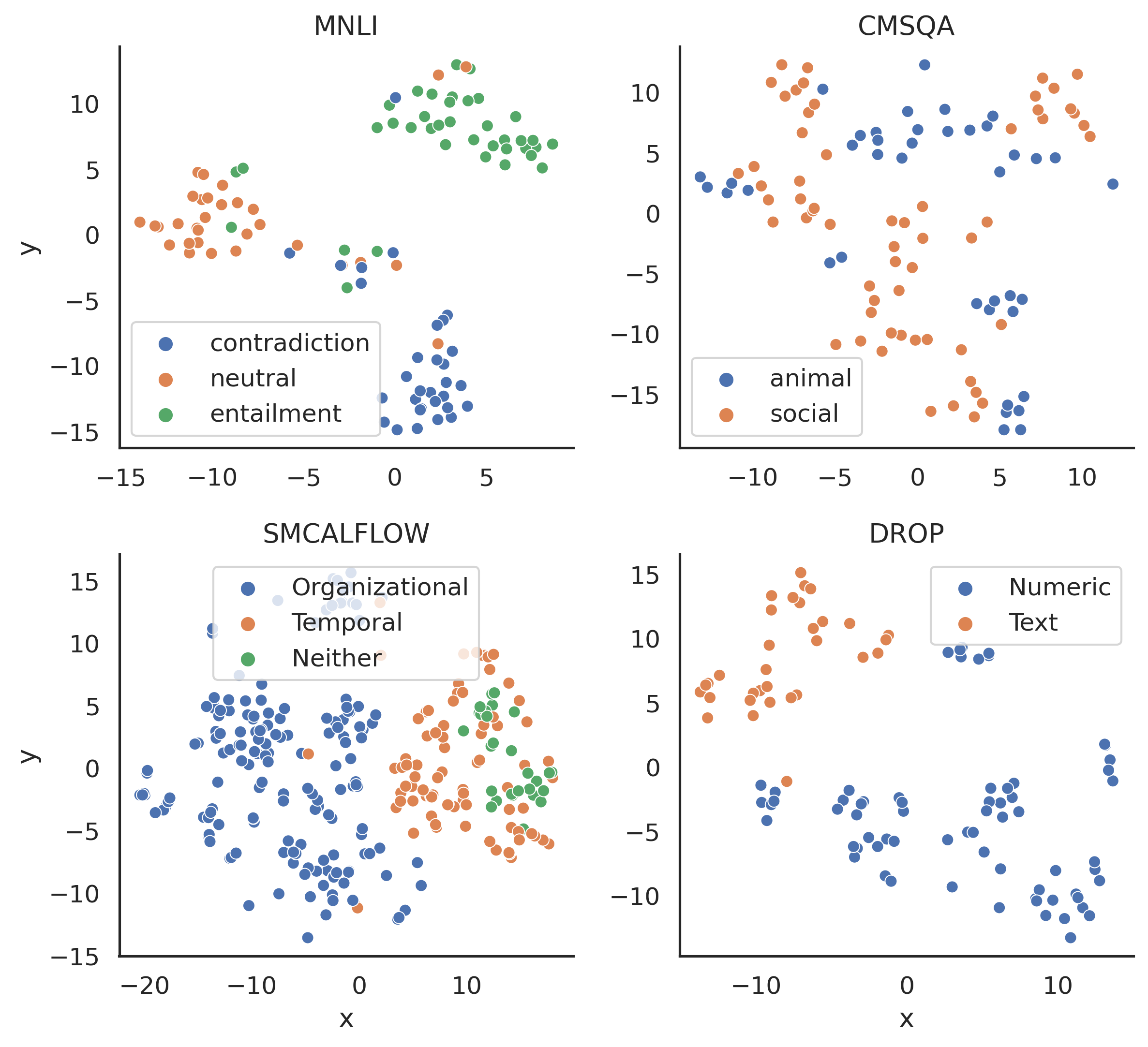}
    \caption{t-SNE visualizations of gist embeddings show that they encode task-specific salient information useful for retrieving informative in-context examples. For \mnli, a classification task, gist embeddings contain information about the class labels. For \cmsqa{}, they encode relevant concepts in the question, \emph{i.e.} whether it's about an animal or an action (\emph{e.g.} "driving car," etc.). For \smcalflow{}, they encode whether the input pertains to organizational hierarchy (\emph{e.g.} \texttt{Who is Bill's manager?}), contains temporal information (\emph{e.g.} \texttt{Book me a dentist appointment before 3pm today}), or neither (\emph{e.g.} \texttt{I need a meeting with Steve}). For \drop{}, they contain information whether the answer is numeric or a textual span.
    }
    \label{fig:qual_projections}
\end{figure}

\begin{resultstablesinglecol}
    \scriptsize
    \setlength{\tabcolsep}{2.5pt}
    \import{tables/tabulars/}{interesting_tasks}
    \caption{%
        Comparison of ICL performance with the performance of the gist models trained on various tasks. Here, the gist model means the full encoder-decoder model with the gist bottleneck. GistScore-based selection can improve ICL performance beyond that of the underlying gist model (\textsc{GM}) itself. In fact, on compositional splits and \gsm{} which requires chain-of-thought reasoning \citep{wei2023chainofthought}, GistScore improves ICL performance even though the gist model itself fails.
        }
    \label{tab:interesting_tasks}
\end{resultstablesinglecol}

We will now analyze the improvements from GistScore. 
For classification tasks, we found ICL accuracy using the different selection methods to be strongly correlated with the precision of selected examples' labels (see Figure \ref{fig:precision_corr}). This suggests that for classification, ICL performance is primarily driven by selection of examples sharing the test input's class label which biases the LLM's prediction. 
GistScore, in particular, does this well because, as shown in Figure \ref{fig:qual_projections}, for classification tasks like \mnli{}, the gist embeddings contain information of the class labels. 
Note that this does not necessarily limit ICL performance for classification tasks---as shown in Figure \ref{fig:precision_lm}, stronger LLMs are less reliant on accurate retrieval and can improve ICL performance beyond selection precision, especially when the selector is inaccurate.

As gist models are trained to perform the tasks, we can also compare against their performance directly. Table \ref{tab:interesting_tasks} shows that ICL with GistScore-based selection can yield performance exceeding that of the underlying gist model itself, especially when using stronger LLMs. This is best exemplified by tasks requiring compositional generalization and chain-of-thought reasoning, a known emergent capability \citep{wei2022emergent}.
This is because, as shown in Figures \ref{fig:qual_projections} and \ref{fig:tsne_projections_2} for \smcalflow{}, \drop{}, and \gsm{}, in these settings, gists can encode abstract task-specific salient aspects \citep{gupta2023coveragebased} useful for selecting informative examples. We share additional analyses of gist embeddings in App. \ref{app:analysis}.

%% file: tables/tabulars/interesting_tasks.tex


\begin{tabular}{clrrrrrrrrr}
\toprule
      & \multirow{2}[4]{*}{\textbf{Method}} & \multicolumn{1}{c}{\multirow{2}[4]{*}{\textbf{\sstfive{}}}} & \multicolumn{1}{c}{\multirow{2}[4]{*}{\textbf{\qnli{}}}} & \multicolumn{1}{c}{\multirow{2}[4]{*}{\textbf{\cmsqa{}}}} & \multicolumn{2}{c}{\textbf{SMC}} & \multicolumn{2}{c}{\textbf{\cogs{}}} & \multicolumn{1}{c}{\multirow{2}[4]{*}{\textbf{GSM}}} & \multicolumn{1}{c}{\multirow{2}[4]{*}{\textbf{\drop{}}}} \\
\cmidrule(lr){6-7} \cmidrule(lr){8-9}
      &       &       &       &       & \multicolumn{1}{c}{\textbf{CG}} & \multicolumn{1}{c}{\textbf{IID}} & \multicolumn{1}{c}{\textbf{CG}} & \multicolumn{1}{c}{\textbf{IID}} &       &  \\
\midrule
      & \textbf{\textsc{GM[F]}} & 53.7  & 85.6  & 64.6  & 0.0   & 64.7  & 45.7  & 99.0  & 0.0   & 32.5 \\
\midrule
\multirow{3}{*}{\rotatebox[origin=c]{90}{\textbf{Neo}}} & \textbf{\rsc{}} & 13.0  & 41.9  & 19.0  & 0.0   & 3.3   & 3.8   & 8.1   & 1.7   & 7.7 \\
      & \textbf{\cossc{}} & 37.9  & 44.0  & 18.1  & 1.1   & 31.6  & 26.0  & 34.7  & 2.0   & 12.6 \\
      & \textbf{\gistftok{1}} & 50.0  & 82.0  & 59.9  & 4.2   & 50.0  & 56.3  & 62.4  & 3.1   & 25.4 \\
\midrule
\multirow{3}{*}{\rotatebox[origin=c]{90}{\textbf{\zephyr{}}}} & \textbf{\rsc{}} & 52.3  & 73.4  & 72.5  & 0.0   & 5.9   & 15.4  & 17.7  & 37.9  & 37.0 \\
      & \textbf{\cossc{}} & 51.2  & 72.1  & 71.6  & 13.4  & 50.8  & 39.7  & 55.4  & 35.9  & 46.3 \\
      & \textbf{\gistftok{1}} & 56.1  & 85.2  & 73.0  & 16.1  & 66.8  & 68.5  & 78.0  & 39.0  & 53.6 \\
\bottomrule
\end{tabular}%

%% file: sections-icml/8_conclusion.tex
\section{Conclusion}
\label{sec:conclusion}

This work presents Example Gisting, a novel approach for training retrievers for in-context learning through supervised fine-tuning of encoder-decoder models with a bottleneck that forces encoding the salient information in inputs into a few tokens. 
We additionally propose GistScore, a novel metric to compare the gist encodings of candidates with the test input. 
Evaluation with a wide range of tasks and LLMs validates the efficacy of our approach by demonstrating the superior performance of our fine-tuned gist models.
Finally, the out-of-the-box generalization of our multi-task trained models enables an improved yet training-free in-context learning pipeline.
Future work could study the efficacy of gisting in other settings that require retrieval, such as retrieval augmented generation.

%% file: sections-icml/9_required.tex
\section*{Impact Statement}

This paper presents a novel approach for retrieving examples for in-context learning with LLMs. While not a specific consequence of our approach, with LLMs, there is always a risk of generating biased, toxic, or non-factual outputs. Further, with in-context learning, the quality and factuality of the retrieved in-context examples also play a role. In turn, these would depend on the bias, toxicity, or factuality of the pool from which the examples are retrieved as well as the retrieval approach. In particular, for our approach, while we don't expect example gisting training to exacerbate these aspects, a gist model could retain any biases of the initial base model as well as the training data used.

%% file: appendices-icml/set_extension.tex
\section{Set Extension}
\label{app:set}

\citet{gupta2023coveragebased} proposed a class of metrics called Coverage Measures for evaluating the relevance of a candidate example $z$ with respect to the test input $x_{\text {test }}$ as a recall of salient aspects with the following form, 
\begin{equation}\label{eq:subscore}
    \operatorname{cover}\left(x_{\text {test }}, z\right)=\sum_{s \in S_{x_{\text{test}}}} \operatorname{c}(s, z)
\end{equation}
where the set of salient aspects $S_{x_{\text{test}}}$ and the coverage of individual aspects $\operatorname{c}(s, z)$ would be defined differently for every metric. Such metrics can be extended to a sub-modular, and hence greedily optimizable, set-level metrics for evaluating sets of examples $Z$ as follows:
\begin{equation}\label{eq:setscore}
    \operatorname{setcov}\left(x_{\text {test }}, Z\right)=\sum_{s \in S_{x_{\text{test}}}} \max_{z \in Z} \operatorname{c}(s, z)
\end{equation}

For $l > 1$ GistScore, as defined in Eq. \ref{eq:gs}, has the form of Eq. \ref{eq:subscore} for $S_{x_{\text{test}}} = \{1, \ldots, L\}$ and $c(s, z) = \frac{1}{l}\max\limits_{j=1, \ldots l} \frac{\mathbf{x}_s^T \mathbf{z}_j}{\lVert\mathbf{x}_s\rVert \lVert\mathbf{z}_j\rVert}$. Thus, its set-extension can be defined as:
\vspace{-5mm}
\begin{equation} \label{eq:setgs}
    \operatorname{Set-GS}_{{l>1}}(x, Z) = \frac{1}{l}\sum\limits_{i=1}^{l} \max_{z \in Z} \max\limits_{j=1, \ldots l} \frac{\mathbf{x}_i^T \mathbf{z}_j}{\lVert\mathbf{x}_i\rVert \lVert\mathbf{z}_j\rVert}
\end{equation}

For $l=1$, GistScore reduces to cosine similarity. Hence, we use \citet{gupta2023coveragebased}'s extension for cosine similarity in this case which assumes $S_{x_{\text{test}}} = \{1, \ldots, d\}$ where $d$ is the embedding size and $c(s, z) = \frac{\mathbf{x}_1[i] \mathbf{z}_1[i]}{\lVert\mathbf{x}_1\rVert \lVert\mathbf{z}_1\rVert}$:
\begin{equation} \label{eq:setgs_gt1}
    \operatorname{Set-GS}_{{l=1}}(x, Z) = \sum\limits_{i=1}^{d} \max_{z \in Z} \frac{\mathbf{x}_1[i] \mathbf{z}_1[i]}{\lVert\mathbf{x}_1\rVert \lVert\mathbf{z}_1\rVert}
\end{equation}

%% file: appendices-icml/datasets.tex
\section{ICL Evaluation}
\label{app:datasets}

\newcommand*{\escape}[1]{\texttt{\textbackslash#1}}

\tightparagraph{Splits} We experiment with 21 datasets spanning 9 task categories. See Table~\ref{tab:ds_summary} for a summary of the datasets used in this work. For all datasets other than \xnli{}, \pawsx{}, \cogs{},  and \smcalflow{}, we use the standard IID splits. For \xnli{} which is a multilingual NLI dataset, we use the German and Russian splits. For \pawsx{} which is a multilingual paraphrase detection dataset, we use the French and Spanish splits. For \cogs{}, we evaluate on the standard IID and compositional generalization evaluation sets. For \smcalflow{} we evaluate on the IID and compositional generalization splits from \citet{yin-etal-2021-compositional} as described below.

\smcalflow{} \citep{andreas-etal-2020-task} is a dataset of task-oriented natural language dialogs about calendars, weather, places, and people paired with executable dataflow programs. \smccs{} \citep{yin-etal-2021-compositional} is a subset of \smcalflow{} containing single-turn dialogs involving two domains (organization structure and calendar event creation), each having its own set of program symbols with two types of test sets: a cross-domain (C) test set containing only instances where both domains appear and meant to test for compositional generalization, and a single-domain (S) test set contains instances with only single-domain for in-distribution evaluation. For compositional evaluation, we use the 32-C split, a few-shot cross-domain split where the training set includes 32 cross-domain examples. For our IID evaluation, following \citet{levy2022diverse}, we use the 8-S split. Additionally, we use the programs with the simplified syntax provided by \cite{meron-2022-simplifying}.

Following prior work \citep{gupta2023coveragebased,rubin-etal-2022-learning,ye2023compositional}, for each split, we use up to 44,000 random instances from the train set as the candidate pool and evaluate on up to 1000 instances from the validation set if available, and the test set otherwise.

\tightparagraph{Templates} Tables \ref{tab:prompts_1}, \ref{tab:prompts_2}, and \ref{tab:prompts_3} contain the textual templates we use to linearize the instances for example selection and ICL.
The ICL prompt is constructed by concatenating the templatized demonstrations and the test instance using \texttt{\escape{n}\escape{n}} as the separator.

\tightparagraph{Evaluation Metric} We report Exact-Match Accuracy for all the Semantic Parsing datasets and Accuracy for the remaining datasets.

\definecolor{codegreen}{rgb}{0,0.6,0}
\definecolor{codegray}{rgb}{0.5,0.5,0.5}
\definecolor{codepurple}{rgb}{0.58,0,0.82}
\definecolor{backcolour}{rgb}{0.95,0.95,0.92}
\lstset{
    language={},
    backgroundcolor=\color{backcolour},
    commentstyle=\color{codegreen},
    keywordstyle=\color{magenta},
    numberstyle=\tiny\color{codegray},
    stringstyle=\color{codepurple},
    basicstyle=\ttfamily\scriptsize,
    literate=%
        {é}{{\'e}}{1}%
        {è}{{\`e}}{1}%
        {à}{{\`a}}{1}%
        {ç}{{\c{c}}}{1}%
        {œ}{{\oe}}{1}%
        {ù}{{\`u}}{1}%
        {É}{{\'E}}{1}%
        {È}{{\`E}}{1}%
        {À}{{\`A}}{1}%
        {Ç}{{\c{C}}}{1}%
        {Œ}{{\OE}}{1}%
        {Ê}{{\^E}}{1}%
        {ê}{{\^e}}{1}%
        {î}{{\^i}}{1}%
        {ô}{{\^o}}{1}%
        {û}{{\^u}}{1}%
        {ë}{{\¨{e}}}1
        {û}{{\^{u}}}1
        {â}{{\^{a}}}1
        {Â}{{\^{A}}}1
        {Î}{{\^{I}}}1
        {á}{{\'a}}1
        {ã}{{\~a}}1
        {é}{{\'e}}1
        {ó}{{\'o}}1
        {í}{{\'i}}1
        {ñ}{{\~n}}1
        {¡}{{!`}}1
        {¿}{{?`}}1
        {ú}{{\'u}}1
        {Í}{{\'I}}1
        {Ó}{{\'O}}1,
    breakatwhitespace=false,
    breaklines=true,
    captionpos=b,
    keepspaces=true,
    numbers=left,
    numbersep=5pt,
    showspaces=false,
    showstringspaces=false,
    showtabs=true,
    tabsize=4,
    aboveskip=-8pt,
    belowskip=-8pt
}

\newcommand*\promptstablelabel{}%
\newcommand*\promptstablecaption{}%
\newenvironment{promptstable}[2]{
    \begingroup
    \setlength{\tabcolsep}{3pt} %
    \def\promptstablelabel{#1}%
    \def\promptstablecaption{#2}%
    \begin{table*}
    \centering
    \small
    \begin{tabular}{p{0.1\linewidth}p{0.42\linewidth}p{0.42\linewidth}}
    \toprule
    Dataset & Selector Example Template & ICL Example Template \\
    \midrule
}{
    \bottomrule
    \end{tabular}
    \caption{\promptstablecaption}
    \label{tab:\promptstablelabel}
    \end{table*}
    \endgroup
}

\begin{promptstable}{prompts_1}{The example templates we use for example selection and in-context learning for the various datasets. See also Tables \ref{tab:prompts_2} and \ref{tab:prompts_3}.}
\smcalflow{} & 
\begin{lstlisting}
Translate this sentence into a logical form representing its meaning: Great , thanks ! I am going to need a meeting with Karen , Jim , and Pam tomorrow before noon .
Logical Form: 
\end{lstlisting} & 
\begin{lstlisting}
Great , thanks ! I am going to need a meeting with Karen , Jim , and Pam tomorrow before noon .	CreateEvent(AND(with_attendee(" Pam "),with_attendee(" Karen "),with_attendee(" Jim "),starts_at(OnDateBeforeTime(date=Tomorrow(),time=Noon()))))
\end{lstlisting} \\
\mtop{} & 
\begin{lstlisting}
Translate this sentence into a logical form representing its meaning: call Nicholas and Natasha
Logical Form: 
\end{lstlisting} & 
\begin{lstlisting}
call Nicholas and Natasha	[IN:CREATE_CALL [SL:CONTACT Nicholas ] [SL:CONTACT Natasha ] ]
\end{lstlisting} \\
\cogs{} & 
\begin{lstlisting}
Translate this sentence into a logical form representing its meaning: Liam hoped that a box was burned by a girl .
Logical Form: 
\end{lstlisting} & 
\begin{lstlisting}
Liam hoped that a box was burned by a girl .	hope ( agent = Liam , ccomp = burn ( theme = box , agent = girl ) )
\end{lstlisting} \\

\qnli{} & 
\begin{lstlisting}
As of that day, the new constitution heralding the Second Republic came into force.
Can we know "What came into force after the new constitution was herald?" given the above sentence (Yes or No)? 
\end{lstlisting} & 
\begin{lstlisting}
Question: What came into force after the new constitution was herald?
Sentence: As of that day, the new constitution heralding the Second Republic came into force.
Answer: Yes
\end{lstlisting} \\
\mnli{} & 
\begin{lstlisting}
Premise: The new rights are nice enough
Does the above premise entail the hypothesis that "Everyone really likes the newest benefits " (Yes, Maybe, or No)?
Answer: 
\end{lstlisting} & 
\begin{lstlisting}
Premise: The new rights are nice enough
Hypothesis: Everyone really likes the newest benefits 
Answer: Maybe
\end{lstlisting} \\
\rte{} & 
\begin{lstlisting}
Dana Reeve, the widow of the actor Christopher Reeve, has died of lung cancer at age 44, according to the Christopher Reeve Foundation.
Based on the above paragraph can we conclude that "Christopher Reeve had an accident." (Yes or No)? 
\end{lstlisting} & 
\begin{lstlisting}
Premise: Dana Reeve, the widow of the actor Christopher Reeve, has died of lung cancer at age 44, according to the Christopher Reeve Foundation.
Hypothesis: Christopher Reeve had an accident.?
Answer: No
\end{lstlisting} \\
\mednli{} & 
\begin{lstlisting}
Premise: No history of blood clots or DVTs, has never had chest pain prior to one week ago.
Is the hypothesis that "Patient has angina." an entailment, contradiction or neutral with respect to the above premise?
Answer: 
\end{lstlisting} & 
\begin{lstlisting}
Premise: No history of blood clots or DVTs, has never had chest pain prior to one week ago.
Hypothesis: Patient has angina.
Answer: Yes
\end{lstlisting} \\
\wanli{} & 
\begin{lstlisting}
Premise: In the past, I have found that there is no point in making a speech unless you have prepared it.
Is the hypothesis that "You should prepare a speech." an entailment, contradiction or neutral with respect to the above premise?
Answer: 
\end{lstlisting} & 
\begin{lstlisting}
Premise: In the past, I have found that there is no point in making a speech unless you have prepared it.
Hypothesis: You should prepare a speech.
Answer: Yes
\end{lstlisting} \\
\xnli{} & 
\begin{lstlisting}
Premise: Et il a dit, maman, je suis à la maison.
Is the hypothesis that "Il a appelé sa mère dès que le bus scolaire l'a déposé." an entailment, contradiction or neutral with respect to the above premise?
Answer: 
\end{lstlisting} & 
\begin{lstlisting}
Premise: Et il a dit, maman, je suis à la maison.
Hypothesis: Il a appelé sa mère dès que le bus scolaire l'a déposé.
Answer: No
\end{lstlisting} \\

\gsm{} & 
\begin{lstlisting}
Give the step-by-step reasoning process and then the final answer.Janet's ducks lay 16 eggs per day. She eats three for breakfast every morning and bakes muffins for her friends every day with four. She sells the remainder at the farmers' market daily for $2 per fresh duck egg. How much in dollars does she make every day at the farmers' market?
\end{lstlisting} & 
\begin{lstlisting}
Question: Janet's ducks lay 16 eggs per day. She eats three for breakfast every morning and bakes muffins for her friends every day with four. She sells the remainder at the farmers' market daily for $2 per fresh duck egg. How much in dollars does she make every day at the farmers' market?
Solution: Janet sells 16 - 3 - 4 = <<16-3-4=9>>9 duck eggs a day.
She makes 9 * 2 = $<<9*2=18>>18 every day at the farmer's market.
#### 18
\end{lstlisting} \\
\agnews{} & 
\begin{lstlisting}
Classify the following news article into one of these categories: World, Sports, Business, Technology.
Fears for T N pension after talks Unions representing workers at Turner   Newall say they are 'disappointed' after talks with stricken parent firm Federal Mogul.
Category: 
\end{lstlisting} & 
\begin{lstlisting}
Article: Fears for T N pension after talks Unions representing workers at Turner   Newall say they are 'disappointed' after talks with stricken parent firm Federal Mogul.
Category: Business
\end{lstlisting} \\
\end{promptstable}

\begin{promptstable}{prompts_2}{The example templates we use for example selection and in-context learning for the various datasets. See also Table \ref{tab:prompts_1} and \ref{tab:prompts_3}.}
\sstfive{} & 
\begin{lstlisting}
Review: in his first stab at the form , jacquot takes a slightly anarchic approach that works only sporadically .
Does the review above see the movie as terrible, bad, OK, good, or great?
Answer: 
\end{lstlisting} & 
\begin{lstlisting}
Review: in his first stab at the form , jacquot takes a slightly anarchic approach that works only sporadically .
Sentiment: OK
\end{lstlisting} \\
\ssttwo{} & 
\begin{lstlisting}
Review: it 's a charming and often affecting journey . 
Is the sentiment of the above review Negative or Positive?
Answer: 
\end{lstlisting} & 
\begin{lstlisting}
Review: it 's a charming and often affecting journey . 
Sentiment: Positive
\end{lstlisting} \\
\rotten{} & 
\begin{lstlisting}
Review: compassionately explores the seemingly irreconcilable situation between conservative christian parents and their estranged gay and lesbian children .
Is the sentiment of the above review Negative or Positive?
Answer: 
\end{lstlisting} & 
\begin{lstlisting}
Review: compassionately explores the seemingly irreconcilable situation between conservative christian parents and their estranged gay and lesbian children .
Sentiment: Positive
\end{lstlisting} \\

\mrpc{} & 
\begin{lstlisting}
Sentence 1: He said the foodservice pie business doesn 't fit the company 's long-term growth strategy .
Sentence 2: " The foodservice pie business does not fit our long-term growth strategy .
Do the above sentences convey the same meaning? Yes or No.
Answer: 
\end{lstlisting} & 
\begin{lstlisting}
Sentence 1: He said the foodservice pie business doesn 't fit the company 's long-term growth strategy .
Sentence 2: " The foodservice pie business does not fit our long-term growth strategy .
Answer: Yes
\end{lstlisting} \\
\paws{} & 
\begin{lstlisting}
Sentence 1: Bradd Crellin represented BARLA Cumbria on a tour of Australia with 6 other players representing Britain , also on a tour of Australia .
Sentence 2: Bradd Crellin also represented BARLA Great Britain on a tour through Australia on a tour through Australia with 6 other players representing Cumbria .
Are these sentences paraphrases of each other? Yes or No.
Answer: 
\end{lstlisting} & 
\begin{lstlisting}
Sentence 1: Bradd Crellin represented BARLA Cumbria on a tour of Australia with 6 other players representing Britain , also on a tour of Australia .
Sentence 2: Bradd Crellin also represented BARLA Great Britain on a tour through Australia on a tour through Australia with 6 other players representing Cumbria .
Answer: No
\end{lstlisting} \\
\qqp{} & 
\begin{lstlisting}
Question 1: Why are African-Americans so beautiful?
Question 2: Why are hispanics so beautiful?
Are Questions 1 and 2 asking the same thing? Yes or No.
Answer: 
\end{lstlisting} & 
\begin{lstlisting}
Question 1: Why are African-Americans so beautiful?
Question 2: Why are hispanics so beautiful?
Answer: No
\end{lstlisting} \\
\pawsx{} & 
\begin{lstlisting}
Sentence 1: El Consejo Shawnee Trail nació de la unión entre el Consejo Four Rivers y el Consejo Audubon.
Sentence 2: El Consejo de caminos de los Shawnee se formó por la fusión del Consejo de Four Rivers y el Consejo de Audubon.
Are the above sentences paraphrases of each other? Yes or No.
Answer: 
\end{lstlisting} & 
\begin{lstlisting}
Sentence 1: El Consejo Shawnee Trail nació de la unión entre el Consejo Four Rivers y el Consejo Audubon.
Sentence 2: El Consejo de caminos de los Shawnee se formó por la fusión del Consejo de Four Rivers y el Consejo de Audubon.
Answer: Yes
\end{lstlisting} \\

\cola{} & 
\begin{lstlisting}
Is the following sentence grammatical (Yes or No)?
The sailors rode the breeze clear of the rocks.
Answer: 
\end{lstlisting} & 
\begin{lstlisting}
Sentence: The sailors rode the breeze clear of the rocks.
Answer: Yes
\end{lstlisting} \\
\tweet{} & 
\begin{lstlisting}
Classify the emotion in the following tweet as one of anger, joy, optimism, or sadness..
Tweet: @user @user Oh, hidden revenge and anger...I rememberthe time,she rebutted you.
Answer: 
\end{lstlisting} & 
\begin{lstlisting}
Tweet: @user @user Oh, hidden revenge and anger...I rememberthe time,she rebutted you.
Answer: A
\end{lstlisting} \\
\cmsqa{} & 
\begin{lstlisting}
Select one of the choices that best answers the following question:
Question: A revolving door is convenient for two direction travel, but it also serves as a security measure at a what?
Option A: bank
Option B: library
Option C: department store
Option D: mall
Option E: new york
Answer: 
\end{lstlisting} & 
\begin{lstlisting}
Question: A revolving door is convenient for two direction travel, but it also serves as a security measure at a what?
Option A: bank
Option B: library
Option C: department store
Option D: mall
Option E: new york
Answer: A
\end{lstlisting} \\
\end{promptstable}

\begin{promptstable}{prompts_3}{The example templates we use for example selection and in-context learning for the various datasets. See also Tables \ref{tab:prompts_1} and \ref{tab:prompts_2}.}
\drop{} & 
\begin{lstlisting}
 Hoping to rebound from their loss to the Patriots, the Raiders stayed at home for a Week 16 duel with the Houston Texans.  Oakland would get the early lead in the first quarter as quarterback JaMarcus Russell completed a 20-yard touchdown pass to rookie wide receiver Chaz Schilens.  The Texans would respond with fullback Vonta Leach getting a 1-yard touchdown run, yet the Raiders would answer with kicker Sebastian Janikowski getting a 33-yard and a 30-yard field goal.  Houston would tie the game in the second quarter with kicker Kris Brown getting a 53-yard and a 24-yard field goal. Oakland would take the lead in the third quarter with wide receiver Johnnie Lee Higgins catching a 29-yard touchdown pass from Russell, followed up by an 80-yard punt return for a touchdown.  The Texans tried to rally in the fourth quarter as Brown nailed a 40-yard field goal, yet the Raiders' defense would shut down any possible attempt.
How many field goals did both teams kick in the first half?
Answer: 
\end{lstlisting} & 
\begin{lstlisting}
Passage:  Hoping to rebound from their loss to the Patriots, the Raiders stayed at home for a Week 16 duel with the Houston Texans.  Oakland would get the early lead in the first quarter as quarterback JaMarcus Russell completed a 20-yard touchdown pass to rookie wide receiver Chaz Schilens.  The Texans would respond with fullback Vonta Leach getting a 1-yard touchdown run, yet the Raiders would answer with kicker Sebastian Janikowski getting a 33-yard and a 30-yard field goal.  Houston would tie the game in the second quarter with kicker Kris Brown getting a 53-yard and a 24-yard field goal. Oakland would take the lead in the third quarter with wide receiver Johnnie Lee Higgins catching a 29-yard touchdown pass from Russell, followed up by an 80-yard punt return for a touchdown.  The Texans tried to rally in the fourth quarter as Brown nailed a 40-yard field goal, yet the Raiders' defense would shut down any possible attempt.
Question: How many field goals did both teams kick in the first half?
Answer: 2
\end{lstlisting} \\
\boolq{} & 
\begin{lstlisting}
Ethanol fuel -- All biomass goes through at least some of these steps: it needs to be grown, collected, dried, fermented, distilled, and burned. All of these steps require resources and an infrastructure. The total amount of energy input into the process compared to the energy released by burning the resulting ethanol fuel is known as the energy balance (or ``energy returned on energy invested''). Figures compiled in a 2007 report by National Geographic Magazine point to modest results for corn ethanol produced in the US: one unit of fossil-fuel energy is required to create 1.3 energy units from the resulting ethanol. The energy balance for sugarcane ethanol produced in Brazil is more favorable, with one unit of fossil-fuel energy required to create 8 from the ethanol. Energy balance estimates are not easily produced, thus numerous such reports have been generated that are contradictory. For instance, a separate survey reports that production of ethanol from sugarcane, which requires a tropical climate to grow productively, returns from 8 to 9 units of energy for each unit expended, as compared to corn, which only returns about 1.34 units of fuel energy for each unit of energy expended. A 2006 University of California Berkeley study, after analyzing six separate studies, concluded that producing ethanol from corn uses much less petroleum than producing gasoline.
does ethanol take more energy make that produces (yes or no)
Answer: 
\end{lstlisting} & 
\begin{lstlisting}
Passage: Ethanol fuel -- All biomass goes through at least some of these steps: it needs to be grown, collected, dried, fermented, distilled, and burned. All of these steps require resources and an infrastructure. The total amount of energy input into the process compared to the energy released by burning the resulting ethanol fuel is known as the energy balance (or ``energy returned on energy invested''). Figures compiled in a 2007 report by National Geographic Magazine point to modest results for corn ethanol produced in the US: one unit of fossil-fuel energy is required to create 1.3 energy units from the resulting ethanol. The energy balance for sugarcane ethanol produced in Brazil is more favorable, with one unit of fossil-fuel energy required to create 8 from the ethanol. Energy balance estimates are not easily produced, thus numerous such reports have been generated that are contradictory. For instance, a separate survey reports that production of ethanol from sugarcane, which requires a tropical climate to grow productively, returns from 8 to 9 units of energy for each unit expended, as compared to corn, which only returns about 1.34 units of fuel energy for each unit of energy expended. A 2006 University of California Berkeley study, after analyzing six separate studies, concluded that producing ethanol from corn uses much less petroleum than producing gasoline.
Question: does ethanol take more energy make that produces
Answer: no
\end{lstlisting} \\
\end{promptstable}

%% file: appendices-icml/training.tex
\section{Training Details}
\label{app:training}

We use encoder-decoder models for both task fine-tuned and multi-task pretrained gist models. This means that after training, we can drop the decoder and only keep the encoder for computing exmaple gists. We experiment with the following different variants of Gist LM-based retrievers:

\tightparagraph{Finetuned Gisting models} (\gistft{}) In this setting, we fine-tune Flan-T5-base \citet{chung2022scaling} models to produce gists of varying lengths on each individual dataset using the procedure described in \S~\ref{sec:example_gisting}. For each dataset, we use the entire train set with instances longer than 500 tokens filtered out for computational efficiency. For early stopping, we compute Rouge-L \citep{lin-2004-rouge} for \drop{} and \gsm{} and Exact-Match Accuracy for the remaining datasets on up to 1000 random instances from the validation set. All training was done with batch size 36 for up to 40000 steps with early stopping with the Adafactor optimizer~\citep{shazeer2018adafactor} and a constant learning rate of 5e-5.

\tightparagraph{Multi-task Pre-trained Gist Model} (\gistmulti{}) For this setting, we train using a large multi-task collection of prompts subsampled from the FLAN 2022 collection \citep{longpre2023flan} of 15M prompts from over 473 datasets and 146 task categories. Specifically, we take zero-shot prompts at most 256 tokens long and further subsample at most 10,000 prompts for every task category.
We use 95\% of this sub-collection for training and 1000 random instances from the remaining 5\% for early stopping with Rouge-L \citep{lin-2004-rouge} as the metric.
To assess effect from varying gist lengths, we train four models that can gist to 1, 3, 6, and 15 tokens.
Each model was trained using the Adafactor optimizer~\citep{shazeer2018adafactor} on an NVIDIA A10G GPU with a batch size of 4 and 64 gradient accumulation steps for an effective batch size of 256. The learning rate was kept constant at 5e-4.

%% file: appendices-icml/all_results.tex
\section{Additional Results}
\label{app:results}

\begin{figure}
    \centering
    \includegraphics[width=\linewidth]{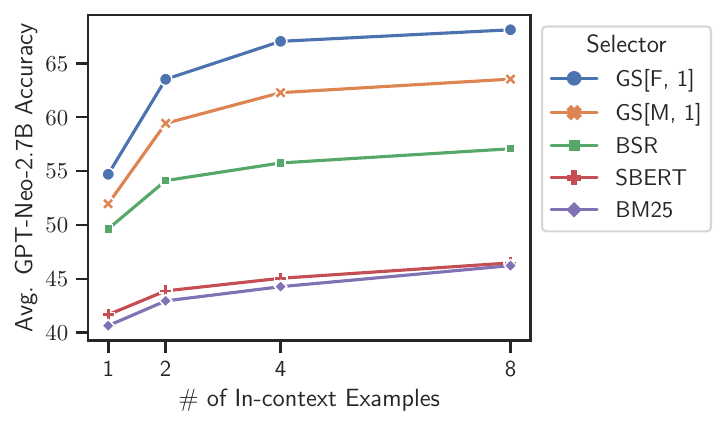}
    \caption{Average ICL performance with \neo{} for varying number of in-context examples. Both \gistft{} and \gistmulti{} are consistently better, and both surpass 8-shot ICL using \cossc{} and \bmsc{} with just 1 example! 
    }
    \label{fig:shots_neo}
\end{figure}

\begin{figure}
    \centering
    \includegraphics[width=\linewidth]{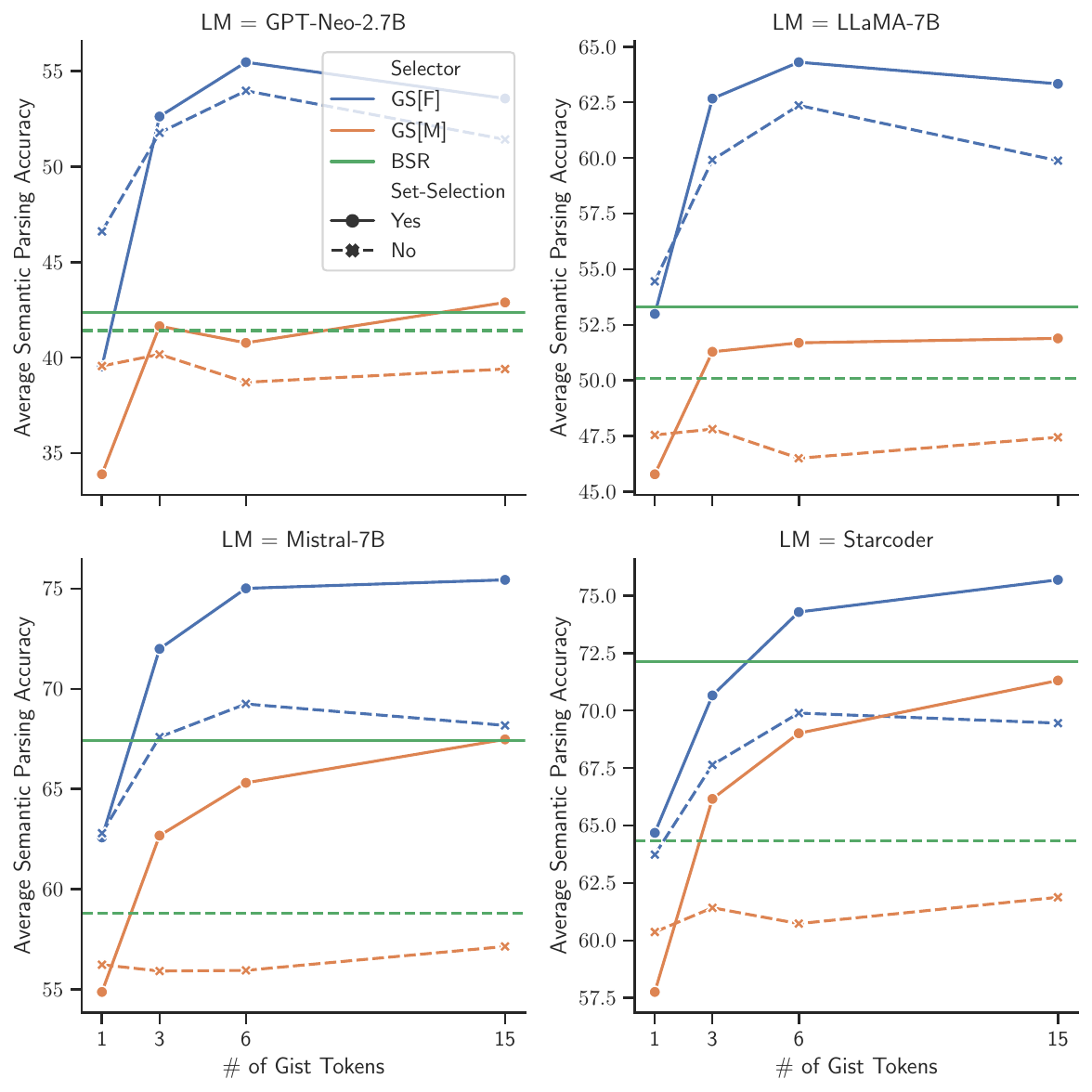}
    \caption{8-shot ICL with different LLMs on semantic parsing datasets using multi-task trained and fine-tuned GistScore with varying number of gist tokens and its set extension.}
    \label{fig:set_all}
\end{figure}

\tightparagraph{Results for GistScore-variations and all baselines} Tables \ref{tab:neo}, \ref{tab:llamaseven}, \ref{tab:llamathirteen}, \ref{tab:mistral}, \ref{tab:zephyr}, \ref{tab:babbage} and \ref{tab:davinci} show 8-shot ICL results for all the datasets with \neo{}, \llamaseven{}, \llamathirteen{}, \mistral{}, \zephyr{}, \babbage{}, and \davinci{}, respectively.

\tightparagraph{Set-selection using \textsc{Set-GS}} Figure \ref{fig:set_all} and Table \ref{tab:set-all} compare performance for different number of gist tokens and set-selection for different LLMs.

\tightparagraph{Varying number of shots} Figure \ref{fig:shots_neo} shows the average ICL performance with \neo{} for varying number of in-context examples.

\tightparagraph{Impact of gist model size} Table \ref{tab:mistral} shows results for GistScore-based selection using a larger multi-task gist model based on \texttt{flan-t5-xl}showing that a stronger gist model can further improve ICL performance.

\tightparagraph{Selection Speeds} Table \ref{tab:speeds} provides the time taken to select 8 ICL examples using various selection methods.

\section{Additional Analyses}
\label{app:analysis}
\begin{figure}
    \centering
    \includegraphics[width=0.8\linewidth]{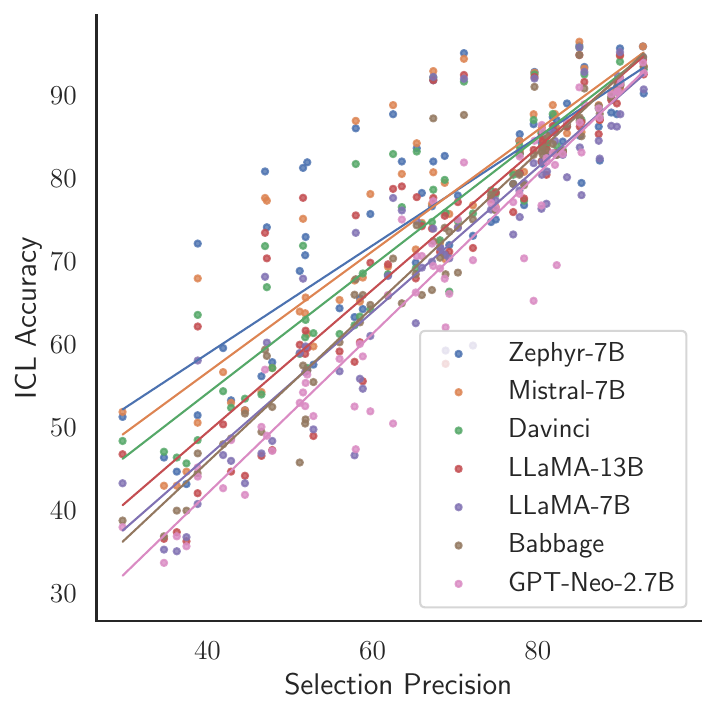}
    \caption{
    ICL accuracy v/s selection precision \emph{i.e.} the fraction of in-context examples with the test label for the various classification datasets with fixed label sets, selectors, and LLMs. While the ICL accuracy of all LLMs improves with more accurate selection, larger LLMs are less reliant on it.
    }
    \label{fig:precision_lm}
\end{figure}

Figure \ref{fig:precision_lm} compares ICL accuracy with selection precision, \textit{i.e.}, the fraction of labels with the test input's label, for classification tasks with fixed label sets and different LLMs.

\begin{figure}
    \centering
    \includegraphics[width=1\linewidth]{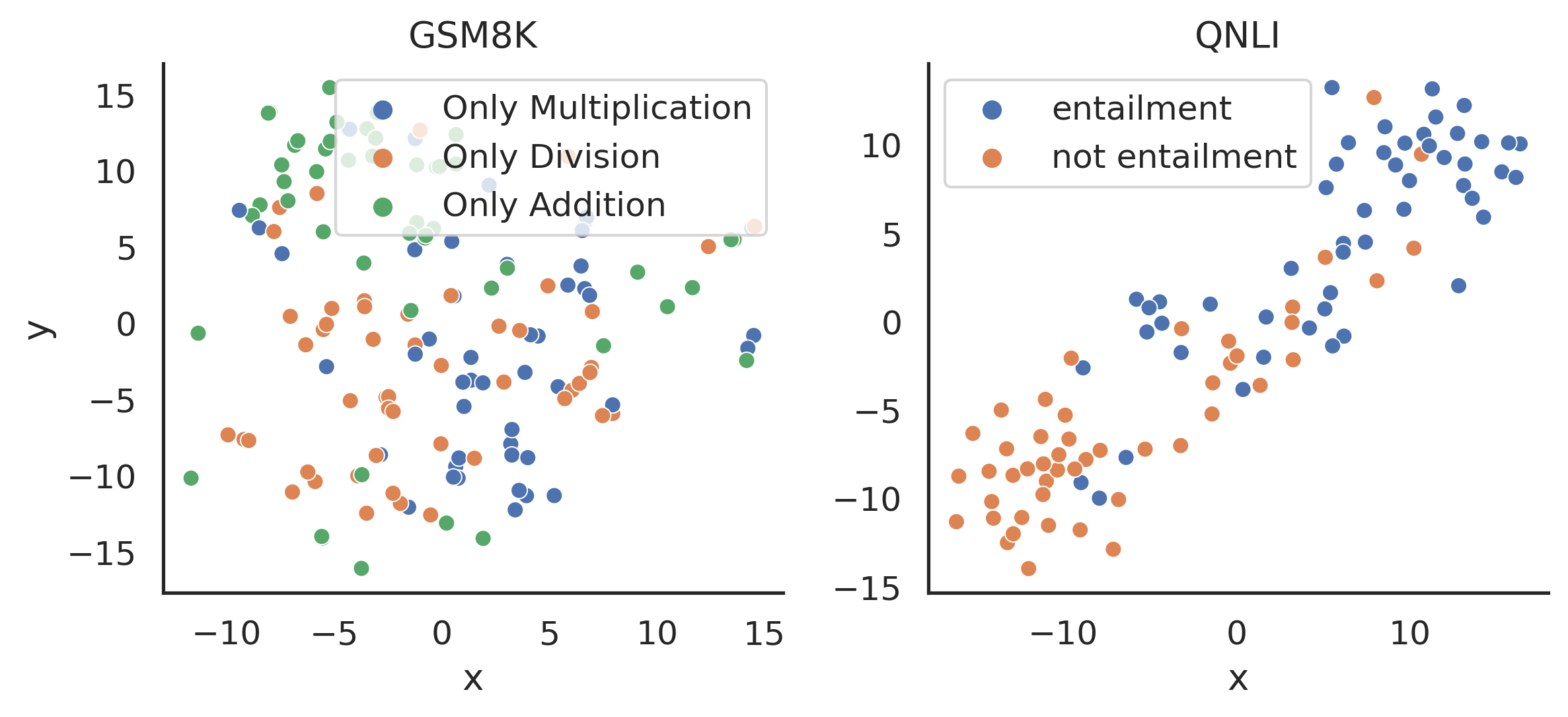}
    \caption{t-SNE Visualizations of gist embeddings for additional datasets. For \qnli{}, gist embeddings encode class labels. For \gsm{}, they encode whether the solution can be obtained by a chain-of-thought reasoning comprising only addition, only multiplication, or only division.}
    \label{fig:tsne_projections_2}
\end{figure}

\begin{figure}
    \centering
    \includegraphics[width=1\linewidth]{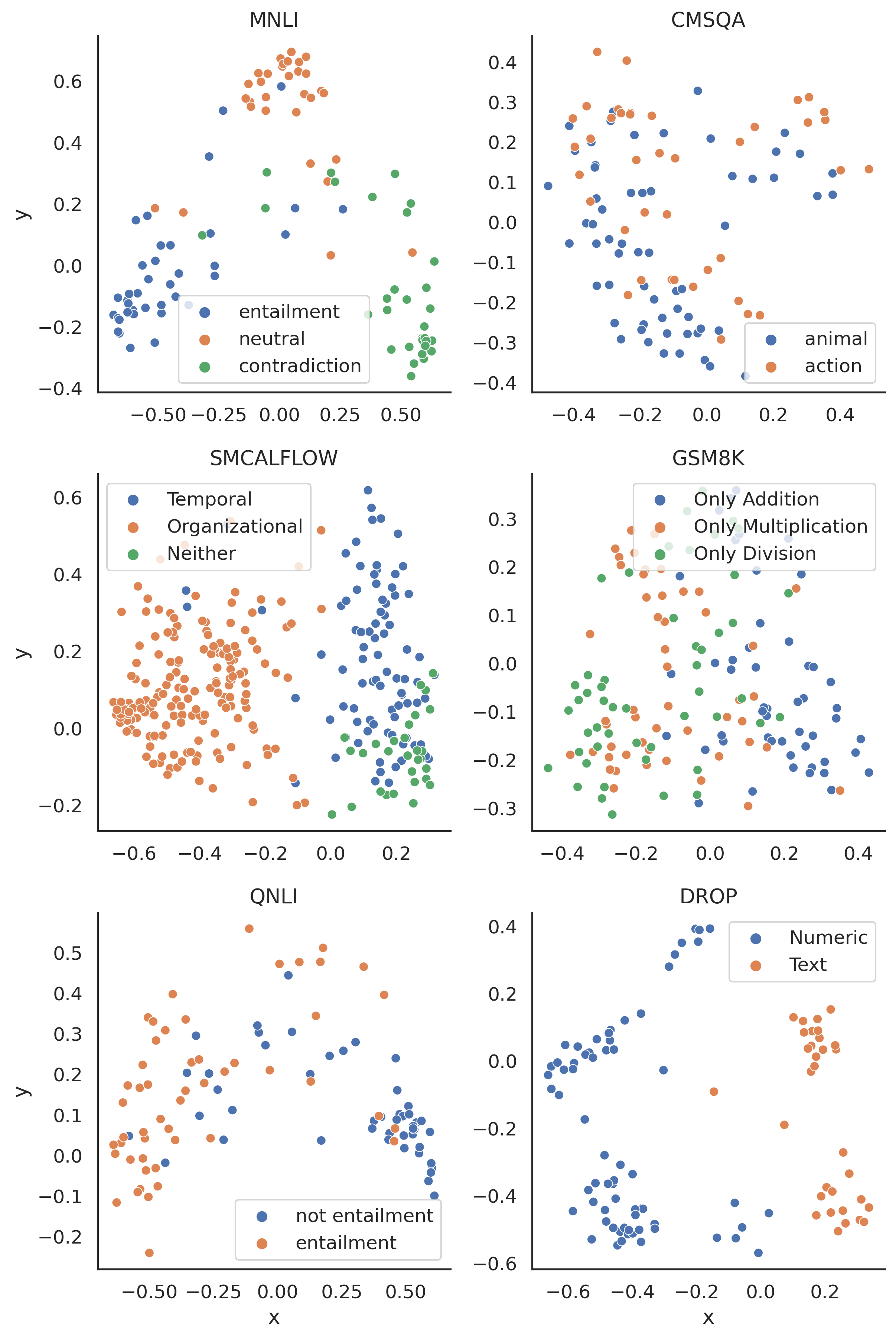}
    \caption{PCA visualizations of gist embeddings show similar results as t-SNE visualization in Figure \ref{fig:qual_projections} and \ref{fig:tsne_projections_2}. Gist embeddings encode task-specific salient information such as class labels (\mnli{}, \qnli{}) or more abstract information aspects (\cmsqa{}, \smcalflow{}, \drop{}, \gsm{}) that help retrieve better in-context examples.}
    \label{fig:pca_projections}
\end{figure}

Figure \ref{fig:tsne_projections_2} shows t-SNE visualizations of salient information in gist embeddings for additional datasets. Figure \ref{fig:pca_projections} shows that the salient aspects seen in t-SNE visualizations in Figures \ref{fig:qual_projections} and \ref{fig:tsne_projections_2} can also be observed in PCA visualizations.

\begin{figure*}[ht!]
    \centering
    \includegraphics[width=1\linewidth]{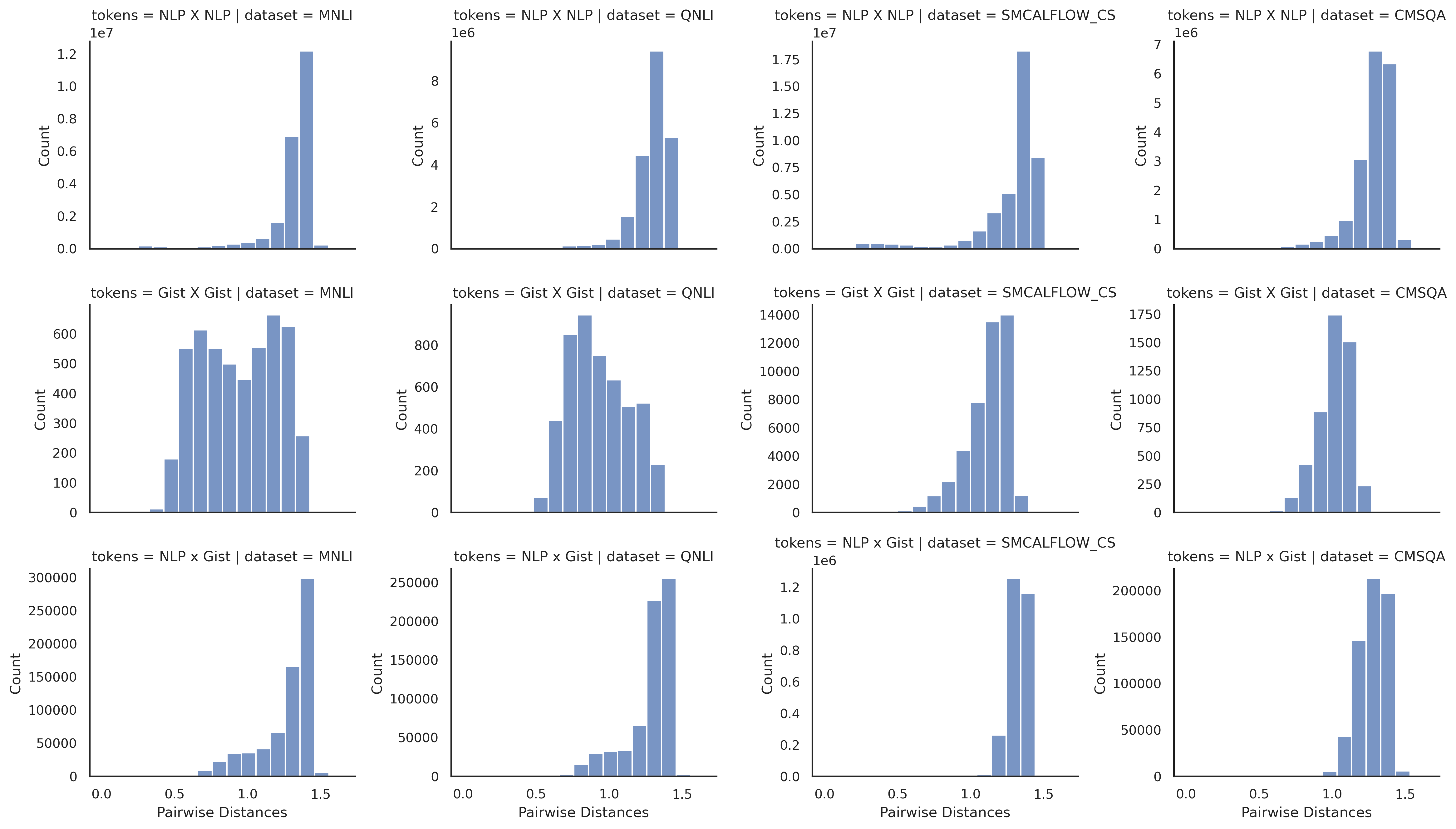}
    \caption{Pairwise Distances between Gist and NLP token activations.}
    \label{fig:qual_pairwise_dists}
\end{figure*}

Figure~\ref{fig:qual_pairwise_dists} qualitatively compares gist token embeddings with ordinary token embeddings through 3 types of pairwise distance distributions: NLP x NLP, Gist x Gist, and NLP x Gist. Clearly, gist tokens are embedded into a different geometry when compared to ordinary language tokens.

\begin{resultstable}
    \setlength{\tabcolsep}{2pt}
    \scriptsize
    \import{tables/tabulars/full_results/}{set-all}
    \caption{
        8-shot ICL results for varying number of gist tokens ($l$) and set-selection for semantic parsing datasets with different LLMs.}
    \label{tab:set-all}
\end{resultstable}

\begin{resultstable}
    \scriptsize
    \import{tables/tabulars/}{sel_times}
    \caption{Time (in ms) to select 8-shots for the various datasets using the different training-free methods. The time for \cossc{} is higher than gisting-based retrieval because our implementation for it does not use FAISS indexing.}
    \label{tab:speeds}
\end{resultstable}

\begin{resultstable}
    \scriptsize
    \import{tables/tabulars/full_results/}{neo}
    \caption{8-shot ICL with \neo{} with independent ranking-based selection. $l$ is the number of gist tokens. {\color[HTML]{990000} Red} highlights datasets or tasks that are held-out from our multi-task training collection. AVG (All) and AVG (Held-out) are average performances on all and only held-out datasets, respectively.}
    \label{tab:neo}
\end{resultstable}

\begin{resultstable}
    \scriptsize
    \import{tables/tabulars/full_results/}{llama-7B}
    \caption{8-shot ICL with \llamaseven{} with independent ranking-based selection. $l$ is the number of gist tokens. {\color[HTML]{990000} Red} highlights datasets or tasks that are held-out from our multi-task training collection. AVG (All) and AVG (Held-out) are average performances on all and only held-out datasets, respectively.}
    \label{tab:llamaseven}
\end{resultstable}

\begin{resultstable}
    \scriptsize
    \import{tables/tabulars/full_results/}{llama-13B}
    \caption{8-shot ICL with \llamathirteen{} with independent ranking-based selection. $l$ is the number of gist tokens. {\color[HTML]{990000} Red} highlights datasets or tasks that are held-out from our multi-task training collection. AVG (All) and AVG (Held-out) are average performances on all and only held-out datasets, respectively.}
    \label{tab:llamathirteen}
\end{resultstable}

\begin{resultstable}
    \scriptsize
    \import{tables/tabulars/full_results/}{mistral}
    \caption{8-shot ICL with \mistral{} with independent ranking-based selection. $l$ is the number of gist tokens. {\color[HTML]{990000} Red} highlights datasets or tasks that are held-out from our multi-task training collection. AVG (All) and AVG (Held-out) are average performances on all and only held-out datasets, respectively.}
    \label{tab:mistral}
\end{resultstable}

\begin{resultstable}
    \scriptsize
    \import{tables/tabulars/full_results/}{zephyr}
    \caption{8-shot ICL with \zephyr{} with independent ranking-based selection. $l$ is the number of gist tokens. {\color[HTML]{990000} Red} highlights datasets or tasks that are held-out from our multi-task training collection. AVG (All) and AVG (Held-out) are average performances on all and only held-out datasets, respectively.}
    \label{tab:zephyr}
\end{resultstable}

\begin{resultstable}
    \scriptsize
    \import{tables/tabulars/full_results/}{babbage}
    \caption{8-shot ICL with \babbage{} with independent ranking-based selection. $l$ is the number of gist tokens. {\color[HTML]{990000} Red} highlights datasets or tasks that are held-out from our multi-task training collection. AVG (All) and AVG (Held-out) are average performances on all and only held-out datasets, respectively.}
    \label{tab:babbage}
\end{resultstable}

\begin{resultstable}
    \scriptsize
    \import{tables/tabulars/full_results/}{davinci}
    \caption{8-shot ICL with \davinci{} with independent ranking-based selection. $l$ is the number of gist tokens. {\color[HTML]{990000} Red} highlights datasets or tasks that are held-out from our multi-task training collection. AVG (All) and AVG (Held-out) are average performances on all and only held-out datasets, respectively.}
    \label{tab:davinci}
\end{resultstable}

%% file: tables/tabulars/full_results/set-all.tex
\begin{tabular}{cccccccccccccccccccc}
\toprule
\multirow{2}[2]{*}{\textbf{LM}} & \multirow{2}[2]{*}{\textbf{Dataset}} & \multirow{2}[2]{*}{\textbf{\bsrsc{}}} & \multirow{2}[2]{*}{\textbf{\setbsrsc{}}} & \multicolumn{4}{c}{\textbf{\gistmulti{}}} & \multicolumn{4}{c}{\textbf{\setgistmulti{}}} & \multicolumn{4}{c}{\textbf{\gistft{}}} & \multicolumn{4}{c}{\textbf{\setgistft{}}} \\
\cmidrule(lr){5-8} \cmidrule(lr){9-12} \cmidrule(lr){13-16} \cmidrule(lr){17-20}
     &      &      &      & \textbf{$l$=1} & \textbf{$l$=3} & \textbf{$l$=6} & \textbf{$l$=15} & \textbf{$l$=1} & \textbf{$l$=3} & \textbf{$l$=6} & \textbf{$l$=15} & \textbf{$l$=1} & \textbf{$l$=3} & \textbf{$l$=6} & \textbf{$l$=15} & \textbf{$l$=1} & \textbf{$l$=3} & \textbf{$l$=6} & \textbf{$l$=15} \\
\midrule
\multirow{6}[4]{*}{\rotatebox[origin=c]{90}{\textbf{\neo}}} & \textbf{SMC CG} & 2.7  & 4.5  & 2.4  & 2    & 2.7  & 3.2  & 1.8  & 5    & 4.8  & 5.1  & 4.2  & 8    & 8.4  & 11.3 & 6.3  & 10.6 & 13   & 13 \\
     & \textbf{SMC IID} & 36.6 & 37   & 33.4 & 35.5 & 34.1 & 35.3 & 30.4 & 36.3 & 34.7 & 37   & 50   & 54.8 & 53.8 & 53.2 & 38.5 & 53.9 & 53.9 & 54.5 \\
     & \textbf{\cogs{} CG} & 52.3 & 53   & 53.3 & 51.6 & 47.9 & 50   & 42.9 & 53.3 & 48.5 & 52   & 56.3 & 64.8 & 68.1 & 64.6 & 50.8 & 65   & 69.3 & 66.6 \\
     & \textbf{\cogs{} IID} & 61.4 & 64.2 & 55.9 & 58.9 & 55.4 & 55.3 & 48.5 & 60.5 & 62.8 & 65.7 & 62.4 & 70.3 & 76.7 & 67.9 & 52.9 & 72.3 & 79.6 & 72.9 \\
     & \textbf{\mtop{}} & 54.1 & 53.1 & 52.8 & 52.9 & 53.4 & 53.2 & 45.9 & 53.2 & 53   & 54.6 & 60.1 & 60.9 & 62.8 & 60.1 & 49.1 & 61.3 & 61.5 & 60.8 \\
\cmidrule{2-20}     & \textbf{AVG} & 41.4  & 42.4  & 39.6  & 40.2  & 38.7  & 39.4  & 33.9  & 41.7  & 40.8  & 42.9  & 46.6  & 51.8  & 54.0  & 51.4  & 39.5  & 52.6  & 55.5  & 53.6 \\
\midrule
\midrule
\multirow{6}[4]{*}{\rotatebox[origin=c]{90}{\textbf{\llamaseven{}}}} & \textbf{SMC CG} & 8.9  & 17.8 & 5.7  & 6.6  & 8    & 7.7  & 9    & 16.7 & 16.9 & 16.4 & 8.7  & 15.8 & 16.1 & 17.6 & 11.6 & 24   & 24.6 & 24.9 \\
     & \textbf{SMC IID} & 51.7 & 53   & 46.8 & 47.7 & 47.7 & 48.9 & 46.4 & 50.6 & 52.9 & 52.4 & 59.8 & 65.4 & 66.8 & 64.4 & 55.7 & 65.6 & 68.1 & 65.9 \\
     & \textbf{\cogs{} CG} & 59.3 & 59.6 & 57.1 & 57   & 52.1 & 54.7 & 53.6 & 59.2 & 55.9 & 56.5 & 63.9 & 70   & 72.6 & 70.5 & 64.5 & 73.2 & 73.7 & 73.1 \\
     & \textbf{\cogs{} IID} & 70.7 & 76   & 69.7 & 68.2 & 66.8 & 64.6 & 65.6 & 70   & 72.8 & 74   & 75.1 & 81   & 87.8 & 80.1 & 75.6 & 83.6 & 87.3 & 84.5 \\
     & \textbf{\mtop{}} & 60   & 60.2 & 58.4 & 59.5 & 57.9 & 61.3 & 54.3 & 59.9 & 60   & 60.1 & 64.7 & 67.3 & 68.5 & 66.8 & 57.5 & 67   & 67.8 & 68.3 \\
\cmidrule{2-20}     & \textbf{AVG} & 50.1  & 53.3  & 47.5  & 47.8  & 46.5  & 47.4  & 45.8  & 51.3  & 51.7  & 51.9  & 54.4  & 59.9  & 62.4  & 59.9  & 53.0  & 62.7  & 64.3  & 63.3 \\
\midrule
\midrule
\multirow{6}[4]{*}{\rotatebox[origin=c]{90}{\textbf{\mistral{}}}} & \textbf{SMC CG} & 17.6 & 49.3 & 13.4 & 13.7 & 15.2 & 17.5 & 23.5 & 35.1 & 45.4 & 46.8 & 17.6 & 27.3 & 27.3 & 32.4 & 28.7 & 45.6 & 49.8 & 56.7 \\
     & \textbf{SMC IID} & 62.4 & 69.8 & 57.6 & 59.7 & 61.9 & 61.3 & 57.9 & 63.7 & 65.3 & 68.7 & 71.5 & 74.8 & 74.6 & 71.5 & 65   & 73.7 & 77.2 & 75.1 \\
     & \textbf{\cogs{} CG} & 65.9 & 66.8 & 64.3 & 62.7 & 61.6 & 63.3 & 59.2 & 65.8 & 64.7 & 68   & 71.7 & 79.1 & 80.7 & 77.6 & 71.6 & 80   & 81.8 & 81.4 \\
     & \textbf{\cogs{} IID} & 80.4 & 82   & 79   & 76.9 & 74.3 & 75.3 & 70.7 & 78.8 & 83   & 84.8 & 81.8 & 86.5 & 90.7 & 86.5 & 82.1 & 88.2 & 92.5 & 90.5 \\
     & \textbf{\mtop{}} & 67.7 & 69.2 & 66.9 & 66.6 & 66.7 & 68.3 & 63.1 & 69.9 & 68.2 & 69.1 & 71.4 & 70.3 & 72.9 & 72.9 & 65.6 & 72.5 & 73.8 & 73.5 \\
\cmidrule{2-20}     & \textbf{AVG} & 58.8  & 67.4  & 56.2  & 55.9  & 55.9  & 57.1  & 54.9  & 62.7  & 65.3  & 67.5  & 62.8  & 67.6  & 69.2  & 68.2  & 62.6  & 72.0  & 75.0  & 75.4 \\
\midrule
\midrule
\multirow{6}[4]{*}{\rotatebox[origin=c]{90}{\textbf{\starcoder{}}}} & \textbf{SMC CG} & 18.6 & 51.4 & 16   & 16.1 & 17.8 & 18.9 & 22.6 & 35.4 & 44.6 & 52.3 & 14.6 & 24.9 & 23.4 & 30.2 & 27.3 & 39.1 & 43.7 & 53.1 \\
     & \textbf{SMC IID} & 65.3 & 69.6 & 58.2 & 60.6 & 59.1 & 63.1 & 55.3 & 63.4 & 65.7 & 69.2 & 69   & 71.6 & 73.3 & 70.7 & 64.5 & 73.4 & 74.8 & 73.7 \\
     & \textbf{\cogs{} CG} & 78   & 77.1 & 70.8 & 72.4 & 71.9 & 70.8 & 64   & 73.2 & 73.4 & 71.6 & 75   & 78.4 & 83.1 & 80.4 & 75.8 & 77.4 & 82.8 & 81.4 \\
     & \textbf{\cogs{} IID} & 91.8 & 92.4 & 88.4 & 88.8 & 86.3 & 87.5 & 81.7 & 88.9 & 92.6 & 91.7 & 89   & 91.8 & 95.6 & 92.6 & 89.9 & 91.3 & 95.6 & 94.7 \\
     & \textbf{\mtop{}} & 68   & 70   & 68.5 & 69.2 & 68.6 & 69.1 & 65.2 & 69.8 & 68.7 & 71.7 & 71   & 71.5 & 74.1 & 73.4 & 65.9 & 72.1 & 74.5 & 75.5 \\
\cmidrule{2-20}     & \textbf{AVG} & 64.3  & 72.1  & 60.4  & 61.4  & 60.7  & 61.9  & 57.8  & 66.1  & 69.0  & 71.3  & 63.7  & 67.6  & 69.9  & 69.5  & 64.7  & 70.7  & 74.3  & 75.7 \\
\bottomrule
\end{tabular}%

%% file: tables/tabulars/sel_times.tex
\begin{tabular}{lccccccccc}
\toprule
\multirow{2}[2]{*}{\textbf{Dataset}} & \multirow{2}[2]{*}{\textbf{\cossc{}}} & \multirow{2}[2]{*}{\textbf{\bmsc{}}} & \multirow{2}[2]{*}{\textbf{\bsrsc{}}} & \multicolumn{4}{c}{\textbf{\gistmulti{}}} & \multicolumn{2}{c}{\textbf{\gistft{}}} \\
\cmidrule(lr){5-8} \cmidrule(lr){9-10}
     &      &      &      & \textbf{$l$=1} & \textbf{$l$=3} & \textbf{$l$=6} & \textbf{$l$=15} & \textbf{$l$=1} & \textbf{$l$=3} \\
\midrule
\smcalflow{} (CG) & 346.84 & 8.72 & 2411.3 & 22.86 & 52.32 & 48.71 & 49.82 & 10.53 & 11.22 \\
\smcalflow{} (IID) & 341.16 & 8.52 & 2418.1 & 23.36 & 56.11 & 26.02 & 27.86 & 13.27 & 12.88 \\
\mtop{} & 169.04 & 5.99 & 723.89 & 23.49 & 55.38 & 54.92 & 59.26 & 10.17 & 10.81 \\
\cogs{} (CG) & 305.8 & 8.62 & 818.34 & 30.41 & 58   & 58.08 & 61.93 & 10.71 & 10.25 \\
\cogs{} (IID) & 297.53 & 8.92 & 816.26 & 23.61 & 56.33 & 51.97 & 62.43 & 11.04 & 11.06 \\
\qnli{} & 1416.9 & 18.21 & 10934 & 1.49 & 2.25 & 3.7  & 7.02 & 1.54 & 2.02 \\
\mnli{} & 1469.9 & 20.71 & 9565.4 & 1.47 & 2.54 & 3.34 & 6.87 & 1.58 & 2.04 \\
\rte{} & 68.81 & 1.01 & 696.73 & 0.79 & 0.8  & 0.69 & 0.92 & 0.57 & 0.83 \\
\wanli{} & 1351 & 19.45 & 6556.2 & 1.45 & 2.17 & 3.81 & 6.89 & 1.53 & 1.98 \\
\xnli{} (de) & 6271 & 51.67 & 28794 & 118.75 & 61.97 & 67.06 & 54.06 & 31.17 & 35.73 \\
\xnli{} (ru) & 5863.8 & 56.89 & 35382 & 125.5 & 56.74 & 56.95 & 62.26 & 35.84 & 30.65 \\
\mednli{} & 285.6 & 4.78 & 3357.4 & 0.74 & 49   & 39.19 & 44.8 & 25.45 & 22 \\
\ssttwo{} & 1119.4 & 22.36 & 3639.3 & 1.52 & 2.31 & 3.36 & 6.99 & 1.53 & 2.14 \\
\sstfive{} & 295.95 & 5.01 & 609.36 & 0.64 & 0.93 & 1.12 & 1.68 & 0.75 & 1.2 \\
\rotten{} & 755.88 & 11.7 & 963.7 & 63.07 & 35.3 & 35.01 & 32.83 & 11.5 & 11.32 \\
\mrpc{} & 89.62 & 1.34 & 255.69 & 0.66 & 0.67 & 0.76 & 1.03 & 0.52 & 0.68 \\
\qqp{} & 1336.2 & 20.13 & 8862.5 & 1.55 & 2.14 & 3.58 & 6.86 & 1.58 & 2.06 \\
\paws{} & 1350.4 & 20.94 & 6712.2 & 1.72 & 2.24 & 3.46 & 6.92 & 1.6  & 2.02 \\
\pawsx{} (es) & 5266.5 & 52.59 & 11698 & 118.93 & 60.19 & 60.61 & 53.79 & 24.65 & 24.95 \\
\pawsx{} (fr) & 5367.5 & 52.03 & 11118 & 114.39 & 59.72 & 56.77 & 53.77 & 24.66 & 25.23 \\
\cmsqa{} & 290.7 & 4.19 & 1124.5 & 0.63 & 0.92 & 1.15 & 1.86 & 0.68 & 0.91 \\
\agnews{} & 2098.1 & 20.78 & 11813 & 1.56 & 2.46 & 3.36 & 6.86 & 1.51 & 1.96 \\
\gsm{} & 138.6 & 3.19 & 1605.9 & 0.61 & 0.84 & 1    & 1.71 & 0.63 & 0.9 \\
\drop{} & 5068.6 & 29.71 & 22340 & 1.51 & 2.23 & 3.24 & 6.75 & 1.48 & 2.01 \\
\boolq{} & 413.46 & 3.75 & 4876 & 0.63 & 0.9  & 1.11 & 1.81 & 0.68 & 0.9 \\
\cola{} & 109.66 & 3.82 & 644.41 & 0.64 & 0.87 & 1.03 & 1.74 & 0.72 & 0.88 \\
\tweet{} (emotion) & 378.42 & 6.57 & 1032 & 104.86 & 40.61 & 45.11 & 38.31 & 11.4 & 11.82 \\
\tweet{} (irony) & 146.37 & 6.22 & 1886 & 90.75 & 34.71 & 46.66 & 39.15 & 11.05 & 11.41 \\
\tweet{} (offensive) & 1201.3 & 20.23 & 4154.5 & 98.87 & 44.79 & 49.41 & 41.29 & 11.75 & 11.42 \\
\tweet{} (sentiment) & 4964.1 & 71.18 & 6870.6 & 124.91 & 61.1 & 60.22 & 62.68 & 24.54 & 25.12 \\
\bottomrule
\end{tabular}%

%% file: tables/tabulars/full_results/neo.tex
\begin{tabular}{lcccccccccccc}
\toprule
\multirow{2}[2]{*}{\textbf{Dataset}} & \multirow{2}[2]{*}{\textbf{\textsc{Rand}}} & \multirow{2}[2]{*}{\textbf{\cossc{}}} & \multirow{2}[2]{*}{\textbf{\bmsc{}}} & \multirow{2}[2]{*}{\textbf{\bsrsc{}}} & \multicolumn{4}{c}{\textbf{\gistmulti{}}} & \multicolumn{2}{c}{\textbf{\gistft{}}} & \multirow{2}[2]{*}{\textbf{\epremph{}}} & \multirow{2}[2]{*}{\textbf{\ceilemph{}}} \\
\cmidrule(lr){6-9} \cmidrule(lr){10-11}
     &      &      &      &      & \textbf{$l$=1} & \textbf{$l$=3} & \textbf{$l$=6} & \textbf{$l$=15} & \textbf{$l$=1} & \textbf{$l$=3} &      &  \\
\midrule
\textcolor[rgb]{ .506,  .086,  .055}{\smcalflow{} (CG)} & 0     & 2.6   & 1.1   & 2.7   & 2.4   & 2     & 2.7   & 3.2   & 4.2   & 8     & 3.6   & 3.8 \\
\textcolor[rgb]{ .506,  .086,  .055}{\smcalflow{} (IID)} & 3.3   & 30.7  & 31.6  & 36.6  & 33.4  & 35.5  & 34.1  & 35.3  & 50    & 54.8  & 54.5  & 59.1 \\
\textcolor[rgb]{ .506,  .086,  .055}{\mtop{}} & 1.3   & 48.4  & 46.4  & 54.1  & 52.8  & 52.9  & 53.4  & 53.2  & 60.1  & 60.9  & 62.2  & 60.5 \\
\textcolor[rgb]{ .506,  .086,  .055}{\cogs{} (CG)} & 3.8   & 25.3  & 26    & 52.3  & 53.3  & 51.6  & 47.9  & 50    & 56.3  & 64.8  &       &  \\
\textcolor[rgb]{ .506,  .086,  .055}{\cogs{} (IID)} & 8.1   & 30.1  & 34.7  & 61.4  & 55.9  & 58.9  & 55.4  & 55.3  & 62.4  & 70.3  &       &  \\
\qnli{} & 54.8  & 56.8  & 56.3  & 82.6  & 86.8  & 85.9  & 85.5  & 85.8  & 91.4  & 93    & 74.9  & 84.2 \\
\mnli{} & 41.9  & 42.2  & 44    & 76.7  & 78.1  & 76.6  & 78.5  & 74.6  & 82    & 81.4  & 66.1  & 71.7 \\
\rte{} & 53.4  & 50.9  & 54.2  & 67.9  & 83    & 77.6  & 81.2  & 73.3  & 81.6  & 81.2  &       &  \\
\textcolor[rgb]{ .506,  .086,  .055}{\wanli{}} & 38.8  & 44.4  & 42.6  & 60    & 58.2  & 53.8  & 53    & 54.8  & 66.2  & 65.4  &       &  \\
\textcolor[rgb]{ .506,  .086,  .055}{\xnli{} (de)} & 33.9  & 36.6  & 33.6  & 41.8  & 58.5  & 56.2  & 58    & 56    & 62    & 62.6  &       &  \\
\textcolor[rgb]{ .506,  .086,  .055}{\xnli{} (ru)} & 32.9  & 34    & 36.8  & 35.6  & 47.1  & 46.5  & 44.3  & 45.7  & 51.3  & 52.5  &       &  \\
\textcolor[rgb]{ .506,  .086,  .055}{\mednli{}} & 41.4  & 54.2  & 56.9  & 70.6  & 69.4  & 71.1  & 69.5  & 70.4  & 82.9  & 83    &       &  \\
\ssttwo{} & 86.9  & 82.6  & 81.9  & 90.9  & 92.1  & 92.4  & 92.5  & 89.6  & 93.9  & 94.3  &       &  \\
\sstfive{} & 13    & 38.9  & 37.9  & 45.1  & 48.4  & 49.3  & 45.9  & 45.3  & 50    & 52.6  & 42.8  & 47 \\
\textcolor[rgb]{ .506,  .086,  .055}{\rotten{}} & 83.1  & 78.1  & 77.2  & 84.5  & 88.9  & 87    & 88.2  & 85.3  & 90.5  & 90.3  &       &  \\
\mrpc{} & 51    & 57.6  & 52.5  & 70.1  & 83.1  & 88    & 84.1  & 75    & 87.3  & 85.3  & 76    & 80.2 \\
\qqp{} & 65.9  & 71.3  & 75    & 86.4  & 85.6  & 85.2  & 85.7  & 84.8  & 86.7  & 88.6  &       &  \\
\paws{} & 48    & 55.2  & 52.5  & 75    & 90.1  & 90.2  & 88.1  & 84.7  & 92.7  & 91.6  &       &  \\
\textcolor[rgb]{ .506,  .086,  .055}{\pawsx{} (es)} & 47.5  & 54.5  & 52.9  & 72.1  & 77.1  & 79.2  & 80.7  & 76    & 88.4  & 86.6  &       &  \\
\textcolor[rgb]{ .506,  .086,  .055}{\pawsx{} (fr)} & 48    & 51.5  & 55.3  & 70.6  & 82.4  & 86.1  & 83    & 81    & 90.4  & 90.2  &       &  \\
\cmsqa{} & 19    & 17.5  & 18.1  & 20.1  & 54.3  & 55.6  & 55    & 44.5  & 59.9  & 57.2  & 36.8  & 37.2 \\
\agnews{} & 76.6  & 89.4  & 89.3  & 89.9  & 91.4  & 90.4  & 90.5  & 90.7  & 92.1  & 92.5  &       &  \\
\gsm{} & 1.7   & 4     & 2     & 2.4   & 3.4   & 1.8   & 3.5   & 3.6   & 3.1   & 3.5   &       &  \\
\drop{} & 7.7   & 12.5  & 12.6  & 10.7  & 18.5  & 18.8  & 19.7  & 18    & 25.4  & 28.7  &       &  \\
\boolq{} & 39.3  & 49.6  & 47.3  & 50.4  & 65.2  & 65    & 66.3  & 59.7  & 69.5  & 66.3  &       &  \\
\cola{} & 60.3  & 64.4  & 64.9  & 69.7  & 76.4  & 75.9  & 74.4  & 70.4  & 80    & 80.3  &       &  \\
\textcolor[rgb]{ .506,  .086,  .055}{\tweet{} (emotion)} & 42.5  & 44.7  & 48.9  & 51.9  & 66    & 69.8  & 64.7  & 59.6  & 70.3  & 70.9  &       &  \\
\textcolor[rgb]{ .506,  .086,  .055}{\tweet{} (offensive)} & 58.8  & 66.5  & 69.1  & 65.9  & 77    & 73.9  & 72.6  & 75.1  & 76.4  & 77.2  &       &  \\
\midrule
AVG (Held-out) & 31.67 & 42.97 & 43.79 & 54.29 & 58.74 & 58.89 & 57.68 & 57.21 & 65.1  & 66.96 &       &  \\
AVG (All) & 37.96 & 46.23 & 46.49 & 57.07 & 63.53 & 63.47 & 62.8  & 60.75 & 68.11 & 69.07 &       &  \\
\bottomrule
\end{tabular}%

%% file: tables/tabulars/full_results/llama-7B.tex
\begin{tabular}{lccccccccccc}
\toprule
\multirow{2}[2]{*}{\textbf{Dataset}} & \multirow{2}[2]{*}{\textbf{\rsc{}}} & \multirow{2}[2]{*}{\textbf{\cossc{}}} & \multirow{2}[2]{*}{\textbf{\bmsc{}}} & \multirow{2}[2]{*}{\textbf{\bsrsc{}}} & \multicolumn{4}{c}{\textbf{\gistmulti{}}} & \multicolumn{2}{c}{\textbf{\gistft{}}} & \multirow{2}[2]{*}{\textbf{\llmremph{}}} \\
\cmidrule(lr){6-9} \cmidrule(lr){10-11}
     &      &      &      &      & \textbf{$l$=1} & \textbf{$l$=3} & \textbf{$l$=6} & \textbf{$l$=15} & \textbf{$l$=1} & \textbf{$l$=3} &  \\
\midrule
\textcolor[rgb]{ .506,  .086,  .055}{\smcalflow{} (CG)} & 0     & 10    & 6.5   & 8.9   & 5.7   & 6.6   & 8     & 7.7   & 8.7   & 15.8  &  \\
\textcolor[rgb]{ .506,  .086,  .055}{\smcalflow{} (IID)} & 6.8   & 45.3  & 45.2  & 51.7  & 46.8  & 47.7  & 47.7  & 48.9  & 59.8  & 65.4  &  \\
\textcolor[rgb]{ .506,  .086,  .055}{\mtop{}} & 3.4   & 54.3  & 53    & 60    & 58.4  & 59.5  & 57.9  & 61.3  & 64.7  & 67.3  &  \\
\textcolor[rgb]{ .506,  .086,  .055}{\cogs{} (CG)} & 13.3  & 29.3  & 32.9  & 59.3  & 57.1  & 57    & 52.1  & 54.7  & 63.9  & 70    &  \\
\textcolor[rgb]{ .506,  .086,  .055}{\cogs{} (IID)} & 10.6  & 35.9  & 42.1  & 70.7  & 69.7  & 68.2  & 66.8  & 64.6  & 75.1  & 81    &  \\
\qnli{} & 51.5  & 57.4  & 56.8  & 75.3  & 80.1  & 82.2  & 81.7  & 79.1  & 87.7  & 90.2  & 69.4 \\
\mnli{} & 54.3  & 56.1  & 58    & 76.3  & 78.5  & 76    & 77.4  & 76.2  & 80.8  & 80.1  & 69.8 \\
\rte{} & 70    & 68.2  & 67.9  & 70.8  & 85.6  & 80.1  & 81.6  & 78.7  & 84.5  & 84.8  & 70.4 \\
\textcolor[rgb]{ .506,  .086,  .055}{\wanli{}} & 45.8  & 47.1  & 46.6  & 55.8  & 56.7  & 55.2  & 53.3  & 52.4  & 62.5  & 63.1  &  \\
\textcolor[rgb]{ .506,  .086,  .055}{\xnli{} (de)} & 40.6  & 37.9  & 35.2  & 43.2  & 54.6  & 54.7  & 53.8  & 52.2  & 59.2  & 61.5  &  \\
\textcolor[rgb]{ .506,  .086,  .055}{\xnli{} (ru)} & 36.5  & 39.7  & 35    & 36.7  & 48.3  & 43.1  & 44    & 45.3  & 49.7  & 52.2  &  \\
\textcolor[rgb]{ .506,  .086,  .055}{\mednli{}} & 60.4  & 69.2  & 68.1  & 74.8  & 73.9  & 75    & 74.6  & 75.3  & 82.8  & 83.6  &  \\
\ssttwo{} & 94.2  & 93.2  & 92    & 95.8  & 95.2  & 94.6  & 94.6  & 94.2  & 94.6  & 94.7  & 93.1 \\
\sstfive{} & 38.4  & 45.2  & 43.2  & 40.7  & 45.9  & 44.8  & 45.1  & 45.6  & 46.8  & 51.2  &  \\
\textcolor[rgb]{ .506,  .086,  .055}{\rotten{}} & 93.1  & 91.3  & 92.2  & 92    & 92.8  & 91.3  & 91.5  & 92.2  & 92.3  & 91.8  &  \\
\mrpc{} & 33.8  & 48.3  & 46.6  & 59.8  & 77.9  & 80.6  & 78.2  & 67.9  & 82.4  & 77.5  & 78.2 \\
\qqp{} & 66.2  & 73.2  & 76.1  & 80.4  & 82    & 80.1  & 79.7  & 80.2  & 83.7  & 84.1  & 83.3 \\
\paws{} & 59.1  & 57.2  & 56.6  & 74    & 86.3  & 88.1  & 87.2  & 80.6  & 90.7  & 89.3  & 57 \\
\textcolor[rgb]{ .506,  .086,  .055}{\pawsx{} (es)} & 57.8  & 59.4  & 58.9  & 69.9  & 73.2  & 76.2  & 75.6  & 72.3  & 84.5  & 81.4  &  \\
\textcolor[rgb]{ .506,  .086,  .055}{\pawsx{} (fr)} & 56.8  & 59.6  & 59.7  & 69.2  & 76.9  & 79.3  & 78.9  & 74.2  & 86.2  & 87.4  &  \\
\cmsqa{} & 39.9  & 26.2  & 29.9  & 30.3  & 60.1  & 63.4  & 62.1  & 49.2  & 63.7  & 60    &  \\
\agnews{} & 85.7  & 88.2  & 86.8  & 88.9  & 90.4  & 90.4  & 90.1  & 88.2  & 90.7  & 92.4  & 93.5 \\
\gsm{} & 11    & 12.4  & 12.3  & 14.3  & 15.6  & 14    & 14.2  & 13.3  & 12.6  & 14.1  &  \\
\drop{} & 24.4  & 28.5  & 27.6  & 27.4  & 32.7  & 32.2  & 31.9  & 31.4  & 36.5  & 39.2  &  \\
\boolq{} & 71.2  & 75.5  & 73.4  & 77.6  & 81.8  & 80.4  & 81.1  & 77.5  & 82.8  & 82.4  & 74.1 \\
\cola{} & 60.1  & 67    & 70.3  & 70.3  & 74.4  & 71.9  & 73.8  & 72.4  & 77.4  & 77.5  &  \\
\textcolor[rgb]{ .506,  .086,  .055}{\tweet{} (emotion)} & 42.8  & 55.6  & 60.2  & 61    & 70.3  & 72.2  & 68.4  & 65.8  & 79.4  & 76.7  &  \\
\textcolor[rgb]{ .506,  .086,  .055}{\tweet{} (offensive)} & 67.6  & 68.7  & 71.6  & 68.2  & 76.2  & 75    & 74.8  & 74.7  & 77.3  & 77    &  \\
\midrule
AVG (Held-out) & 38.25 & 50.24 & 50.51 & 58.67 & 61.47 & 61.5  & 60.53 & 60.11 & 67.58 & 69.59 &  \\
AVG (All) & 46.26 & 53.57 & 53.74 & 60.83 & 65.97 & 65.71 & 65.22 & 63.43 & 70.04 & 71.13 &  \\
\bottomrule
\end{tabular}%

%% file: tables/tabulars/full_results/llama-13B.tex
\begin{tabular}{lcccccccccc}
\toprule
\multirow{2}[2]{*}{\textbf{Dataset}} & \multirow{2}[2]{*}{\textbf{\rsc{}}} & \multirow{2}[2]{*}{\textbf{\cossc{}}} & \multirow{2}[2]{*}{\textbf{\bmsc{}}} & \multirow{2}[2]{*}{\textbf{\bsrsc{}}} & \multicolumn{4}{c}{\textbf{\gistmulti{}}} & \multicolumn{2}{c}{\textbf{\gistft{}}} \\
\cmidrule(lr){6-9} \cmidrule(lr){10-11}
     &      &      &      &      & \textbf{$l$=1} & \textbf{$l$=3} & \textbf{$l$=6} & \textbf{$l$=15} & \textbf{$l$=1} & \textbf{$l$=3} \\
\midrule
\textcolor[rgb]{ .506,  .086,  .055}{\smcalflow{} (CG)} & 0     & 12.4  & 9.5   & 12.7  & 10.1  & 8.7   & 8.7   & 11.5  & 10.6  & 19.9 \\
\textcolor[rgb]{ .506,  .086,  .055}{\smcalflow{} (IID)} & 15.3  & 48.8  & 49.4  & 57.4  & 50.3  & 55    & 52.4  & 53.5  & 60.7  & 62.8 \\
\textcolor[rgb]{ .506,  .086,  .055}{\mtop{}} & 3.9   & 59.7  & 56.5  & 63.4  & 61.4  & 62.4  & 61.2  & 65    & 68.8  & 68.7 \\
\textcolor[rgb]{ .506,  .086,  .055}{\cogs{} (CG)} & 14.8  & 31.7  & 35.5  & 60.2  & 56.8  & 57.1  & 53    & 56.3  & 66.5  & 70.5 \\
\textcolor[rgb]{ .506,  .086,  .055}{\cogs{} (IID)} & 16.3  & 41.3  & 48.5  & 71.7  & 69.8  & 71.1  & 68.6  & 68.9  & 76.9  & 82.2 \\
\qnli{} & 56.7  & 59.7  & 59.5  & 80.6  & 86.2  & 86.1  & 85.4  & 85.8  & 91.2  & 92.6 \\
\mnli{} & 50.3  & 61.9  & 62.1  & 82    & 80.6  & 80.4  & 80.6  & 78.4  & 83.4  & 81.4 \\
\rte{} & 76.5  & 73.3  & 77.6  & 75.5  & 86.3  & 82.7  & 83    & 80.5  & 85.6  & 84.5 \\
\textcolor[rgb]{ .506,  .086,  .055}{\wanli{}} & 44    & 50    & 50.3  & 60.2  & 59.1  & 58.1  & 56.4  & 59.5  & 67.9  & 67.1 \\
\textcolor[rgb]{ .506,  .086,  .055}{\xnli{} (de)} & 36.1  & 40.6  & 36.5  & 44.1  & 55.5  & 57.8  & 56.6  & 53.6  & 57.6  & 59.9 \\
\textcolor[rgb]{ .506,  .086,  .055}{\xnli{} (ru)} & 34.5  & 38    & 37.3  & 36.2  & 47.2  & 44.9  & 46.6  & 48.1  & 48.9  & 53.9 \\
\textcolor[rgb]{ .506,  .086,  .055}{\mednli{}} & 54.5  & 71.9  & 73.4  & 77.7  & 77.6  & 78    & 78.4  & 77.9  & 83.2  & 84.6 \\
\ssttwo{} & 93.5  & 93    & 92.4  & 94.8  & 94.8  & 94.6  & 94.4  & 93.3  & 94.3  & 94.7 \\
\sstfive{} & 40    & 46.2  & 46.7  & 42    & 44.6  & 43.5  & 46.8  & 43.2  & 46.5  & 48 \\
\textcolor[rgb]{ .506,  .086,  .055}{\rotten{}} & 87.1  & 91.6  & 91.8  & 92.2  & 91.6  & 92.1  & 91.8  & 92.9  & 91.7  & 91.5 \\
\mrpc{} & 70.6  & 62.7  & 57.8  & 71.6  & 86.8  & 88    & 85.5  & 77.2  & 87    & 86 \\
\qqp{} & 66.8  & 77.2  & 79    & 85.1  & 84.4  & 83.4  & 84.2  & 84.2  & 86.2  & 87.4 \\
\paws{} & 59.7  & 58.5  & 58.8  & 77.1  & 89.4  & 90.2  & 89.3  & 85.3  & 92.5  & 91.7 \\
\textcolor[rgb]{ .506,  .086,  .055}{\pawsx{} (es)} & 60.2  & 60.5  & 59.9  & 73.9  & 75.9  & 78.4  & 77.8  & 75.1  & 85.3  & 83.1 \\
\textcolor[rgb]{ .506,  .086,  .055}{\pawsx{} (fr)} & 63.4  & 63.5  & 61.6  & 74.2  & 80.4  & 84.6  & 82.4  & 79.6  & 89    & 90 \\
\cmsqa{} & 51.4  & 41    & 44    & 42.2  & 64.7  & 68.4  & 67.4  & 60.4  & 64.9  & 62.2 \\
\agnews{} & 83.9  & 91.6  & 91.2  & 91.3  & 92.9  & 92.8  & 92.7  & 91.2  & 93.4  & 93.9 \\
\gsm{} & 15.4  & 16.4  & 16.7  & 19.4  & 16.8  & 18.2  & 18.1  & 18.6  & 18.9  & 17.3 \\
\drop{} & 31.1  & 33.5  & 32.9  & 33.2  & 37.3  & 36.7  & 38.4  & 36.7  & 42.7  & 42.9 \\
\boolq{} & 63.4  & 77    & 75.5  & 78.7  & 83.4  & 82.7  & 82.6  & 80.3  & 83    & 82.7 \\
\cola{} & 58.9  & 65.4  & 71    & 72.4  & 76    & 74.5  & 76.8  & 72.9  & 80.1  & 79.5 \\
\textcolor[rgb]{ .506,  .086,  .055}{\tweet{} (emotion)} & 55.3  & 67.9  & 70.3  & 69.8  & 71.1  & 73    & 74.6  & 74.1  & 77.5  & 78.6 \\
\textcolor[rgb]{ .506,  .086,  .055}{\tweet{} (offensive)} & 66.7  & 69.9  & 71.1  & 69.6  & 77.6  & 76    & 75.7  & 75.5  & 78.3  & 78.3 \\
\midrule
AVG (Held-out) & 39.44 & 53.41 & 53.69 & 61.66 & 63.17 & 64.09 & 63.16 & 63.68 & 68.78 & 70.79 \\
AVG (All) & 48.94 & 57.33 & 57.74 & 64.61 & 68.16 & 68.55 & 68.19 & 67.13 & 71.88 & 72.71 \\
\bottomrule
\end{tabular}%

%% file: tables/tabulars/full_results/mistral.tex
\begin{tabular}{lccccccccccc}
\toprule
\multirow{2}[2]{*}{\textbf{Dataset}} & \multirow{2}[2]{*}{\textbf{\rsc{}}} & \multirow{2}[2]{*}{\textbf{\bmsc{}}} & \multirow{2}[2]{*}{\textbf{\cossc{}}} & \multirow{2}[2]{*}{\textbf{\bsrsc{}}} & \multicolumn{4}{c}{\textbf{\textsc{GS[M, Large]}}} & \multicolumn{2}{c}{\textbf{\gistft{}}} & \textbf{\textsc{GS[M, XL]}} \\
      &       &       &       &       & \boldmath{}\textbf{$l = 1$}\unboldmath{} & \boldmath{}\textbf{$l = 3$}\unboldmath{} & \boldmath{}\textbf{$l = 6$}\unboldmath{} & \boldmath{}\textbf{$l = 15$}\unboldmath{} & \boldmath{}\textbf{$l = 1$}\unboldmath{} & \boldmath{}\textbf{$l = 3$}\unboldmath{} & \boldmath{}\textbf{$l = 1$}\unboldmath{} \\
\midrule
\textcolor[rgb]{ .506,  .086,  .055}{\smcalflow{} (CG)} & 0     & 21.6  & 15.7  & 17.6  & 13.4  & 13.7  & 15.2  & 17.5  & 17.6  & 27.3  & 18.1 \\
\textcolor[rgb]{ .506,  .086,  .055}{\smcalflow{} (IID)} & 13.7  & 55.7  & 57.1  & 62.4  & 57.6  & 59.7  & 61.9  & 61.3  & 71.5  & 74.8  & 64 \\
\textcolor[rgb]{ .506,  .086,  .055}{\mtop{}} & 7     & 63.5  & 60.2  & 67.7  & 66.9  & 66.6  & 66.7  & 68.3  & 71.4  & 70.3  & 67.1 \\
\textcolor[rgb]{ .506,  .086,  .055}{\cogs{} (CG)} & 14.2  & 35.2  & 42.7  & 65.9  & 64.3  & 62.7  & 61.6  & 63.3  & 71.7  & 79.1  & 60.5 \\
\textcolor[rgb]{ .506,  .086,  .055}{\cogs{} (IID)} & 18.4  & 48    & 58.7  & 80.4  & 79    & 76.9  & 74.3  & 75.3  & 81.8  & 86.5  & 75.9 \\
\qnli{} & 56.4  & 62.8  & 61.2  & 83.3  & 85.4  & 86.4  & 86.9  & 85.8  & 90.6  & 92.3  & 87.9 \\
\mnli{} & 62    & 67.6  & 67.9  & 85.6  & 84.5  & 82.2  & 82.2  & 82.5  & 85    & 85.7  & 85.7 \\
\rte{} & 80.1  & 77.3  & 75.1  & 79.4  & 88.8  & 84.5  & 83.4  & 83.8  & 87.7  & 84.8  & 88.4 \\
\textcolor[rgb]{ .506,  .086,  .055}{\wanli{}} & 54.5  & 56.3  & 56.6  & 65.1  & 65.3  & 60.1  & 63    & 61.8  & 71.4  & 71.3  & 65.7 \\
\textcolor[rgb]{ .506,  .086,  .055}{\xnli{} (de)} & 35.1  & 46.3  & 42.9  & 52    & 68    & 66.9  & 68.1  & 63.8  & 70.2  & 70.9  & 71.1 \\
\textcolor[rgb]{ .506,  .086,  .055}{\xnli{} (ru)} & 33.4  & 42.8  & 42.9  & 44.6  & 57.1  & 55.5  & 55.1  & 54.3  & 59.7  & 58.3  & 60.4 \\
\textcolor[rgb]{ .506,  .086,  .055}{\mednli{}} & 75.4  & 78.7  & 77.6  & 84.2  & 80.7  & 82    & 83.3  & 82.5  & 83.1  & 85    & 83.5 \\
\ssttwo{} & 95.5  & 94.5  & 94.4  & 96.4  & 94.7  & 94.7  & 96    & 95    & 95.9  & 95.6  & 94.8 \\
\sstfive{} & 51.1  & 51.1  & 51.8  & 50.5  & 52.9  & 53.2  & 52.7  & 52.7  & 54.2  & 55.4  & 53.6 \\
\textcolor[rgb]{ .506,  .086,  .055}{\rotten{}} & 93.3  & 91.9  & 92.9  & 92.7  & 93.2  & 92.5  & 92.5  & 93.5  & 90.7  & 91.8  & 92.6 \\
\mrpc{} & 72.8  & 70.6  & 67.6  & 76.7  & 85.5  & 88    & 84.6  & 79.7  & 87    & 87    & 90.4 \\
\qqp{} & 73.8  & 78.5  & 80.5  & 86.1  & 84.8  & 84.4  & 84.3  & 85.5  & 86.9  & 88.5  & 85.1 \\
\paws{} & 71.2  & 60.8  & 63.7  & 74.1  & 90.5  & 91.3  & 90.4  & 88.1  & 93.5  & 92.5  & 92.5 \\
\textcolor[rgb]{ .506,  .086,  .055}{\pawsx{} (es)} & 68.8  & 63.3  & 63.9  & 76.9  & 80.7  & 82.2  & 82.2  & 77.8  & 88.8  & 87.2  & 86.3 \\
\textcolor[rgb]{ .506,  .086,  .055}{\pawsx{} (fr)} & 71.7  & 63.8  & 65.6  & 74.6  & 83.9  & 86.4  & 84.1  & 82.5  & 90.8  & 90.7  & 86.8 \\
\cmsqa{} & 73.5  & 67.6  & 70.6  & 69    & 75.1  & 76.4  & 76.5  & 72.7  & 74.2  & 73.3  & 77.8 \\
\agnews{} & 88.3  & 93.4  & 93.2  & 93.1  & 94.6  & 94.4  & 93.7  & 92.9  & 93.8  & 94.4  & 94.5 \\
\gsm{} & 34.8  & 37.3  & 37    & 40    & 37.9  & 37.6  & 39.4  & 38.7  & 38.5  & 40.3  & 42.2 \\
\drop{} & 41.1  & 48.3  & 48.2  & 48.4  & 56    & 54.8  & 53.9  & 54.8  & 58.5  & 59.2  & 56.2 \\
\boolq{} & 86.4  & 87.3  & 86.9  & 88.8  & 87.7  & 88.9  & 87.2  & 87.9  & 86    & 86.5  & 89.1 \\
\cola{} & 82.1  & 82.2  & 83.1  & 82.2  & 81.8  & 80.3  & 81.1  & 82    & 83    & 83.2  & 83 \\
\textcolor[rgb]{ .506,  .086,  .055}{\tweet{} (emotion)} & 59.1  & 75.4  & 77.3  & 78.1  & 75.7  & 78.3  & 78.6  & 77.5  & 80.7  & 82.9  & 78.9 \\
\textcolor[rgb]{ .506,  .086,  .055}{\tweet{} (offensive)} & 65.7  & 69.3  & 72.2  & 69.3  & 77.4  & 75.1  & 75.4  & 74.3  & 76.5  & 76.9  & 78.8 \\
\midrule
AVG (Held-out) & 43.59 & 57.99 & 59.02 & 66.54 & 68.8  & 68.47 & 68.71 & 68.12 & 73.28 & 75.21 & 70.69 \\
AVG (All) & 56.41 & 63.97 & 64.55 & 70.9  & 73.69 & 73.42 & 73.37 & 72.71 & 76.45 & 77.56 & 75.39 \\
\bottomrule
\end{tabular}%

%% file: tables/tabulars/full_results/zephyr.tex
\begin{tabular}{lcccccccccc}
\toprule
\multirow{2}[2]{*}{\textbf{Dataset}} & \multirow{2}[2]{*}{\textbf{\rsc{}}} & \multirow{2}[2]{*}{\textbf{\cossc{}}} & \multirow{2}[2]{*}{\textbf{\bmsc{}}} & \multirow{2}[2]{*}{\textbf{\bsrsc{}}} & \multicolumn{4}{c}{\textbf{\gistmulti{}}} & \multicolumn{2}{c}{\textbf{\gistft{}}} \\
\cmidrule(lr){6-9} \cmidrule(lr){10-11}
     &      &      &      &      & \textbf{$l$=1} & \textbf{$l$=3} & \textbf{$l$=6} & \textbf{$l$=15} & \textbf{$l$=1} & \textbf{$l$=3} \\
\midrule
\textcolor[rgb]{ .506,  .086,  .055}{\smcalflow{} (CG)} & 0     & 19    & 13.4  & 15.8  & 11.8  & 12.1  & 13.3  & 15.1  & 16.1  & 23.7 \\
\textcolor[rgb]{ .506,  .086,  .055}{\smcalflow{} (IID)} & 5.9   & 51.1  & 50.8  & 56.6  & 51.1  & 53.6  & 57.6  & 59.7  & 66.8  & 69.3 \\
\textcolor[rgb]{ .506,  .086,  .055}{\mtop{}} & 4.7   & 59    & 54    & 61.3  & 61    & 61.1  & 59.8  & 62.3  & 67    & 65.9 \\
\textcolor[rgb]{ .506,  .086,  .055}{\cogs{} (CG)} & 15.4  & 33.8  & 39.7  & 63.3  & 61.4  & 59.6  & 59.3  & 61.8  & 68.5  & 76.1 \\
\textcolor[rgb]{ .506,  .086,  .055}{\cogs{} (IID)} & 17.7  & 46.6  & 55.4  & 77.4  & 74.7  & 72    & 72.4  & 70.9  & 78    & 83 \\
\qnli{} & 81.7  & 81.3  & 81.9  & 85.3  & 89    & 87.8  & 88.4  & 88.8  & 91.6  & 92.3 \\
\mnli{} & 73.4  & 72.5  & 72.1  & 84.3  & 84.5  & 83.3  & 83.7  & 83.7  & 85.2  & 84.5 \\
\rte{} & 80.5  & 81.6  & 81.2  & 82.7  & 87.4  & 83.4  & 85.2  & 85.6  & 86.3  & 85.2 \\
\textcolor[rgb]{ .506,  .086,  .055}{\wanli{}} & 50.5  & 58.8  & 59.5  & 65.5  & 64.3  & 62.1  & 63.4  & 63.4  & 69.8  & 69.3 \\
\textcolor[rgb]{ .506,  .086,  .055}{\xnli{} (de)} & 42.5  & 45.9  & 46.3  & 52    & 64.2  & 64.6  & 64.1  & 61.5  & 70.8  & 69.3 \\
\textcolor[rgb]{ .506,  .086,  .055}{\xnli{} (ru)} & 42.8  & 44.7  & 44.6  & 43.1  & 57.8  & 55.4  & 53.1  & 53.5  & 57.5  & 58.9 \\
\textcolor[rgb]{ .506,  .086,  .055}{\mednli{}} & 76.3  & 80    & 80.8  & 83.6  & 82    & 83.9  & 83.8  & 82.3  & 84.4  & 85.3 \\
\ssttwo{} & 95.6  & 94.8  & 95.1  & 96    & 95.6  & 96.1  & 96.1  & 96.1  & 95.9  & 96.1 \\
\sstfive{} & 52.3  & 51.6  & 51.2  & 51.4  & 53.2  & 52.8  & 52.7  & 53.9  & 56.1  & 55.2 \\
\textcolor[rgb]{ .506,  .086,  .055}{\rotten{}} & 92.5  & 91.1  & 91.8  & 92.8  & 93.4  & 93.3  & 92.9  & 93.3  & 91.3  & 92.4 \\
\mrpc{} & 74.3  & 67.9  & 63.2  & 73    & 79.4  & 83.3  & 80.6  & 74.3  & 82.1  & 82.4 \\
\qqp{} & 80.2  & 80    & 82    & 82    & 81.7  & 82.3  & 81.5  & 83.5  & 85.1  & 84.6 \\
\paws{} & 71.7  & 68.5  & 70.7  & 77.9  & 87.9  & 85.8  & 85.7  & 84.7  & 90.2  & 88.9 \\
\textcolor[rgb]{ .506,  .086,  .055}{\pawsx{} (es)} & 73.5  & 69.1  & 68.8  & 76.6  & 79.3  & 81.4  & 81.7  & 77.7  & 86.2  & 86 \\
\textcolor[rgb]{ .506,  .086,  .055}{\pawsx{} (fr)} & 72.9  & 69.9  & 72.9  & 78.2  & 82.6  & 82.9  & 81.8  & 80.9  & 87.7  & 87.4 \\
\cmsqa{} & 72.5  & 67.7  & 71.6  & 68.9  & 71.8  & 74.1  & 72.9  & 71.5  & 73    & 72.2 \\
\agnews{} & 87.8  & 93.3  & 92.6  & 93.1  & 93.8  & 93.5  & 93.9  & 92.3  & 92.6  & 93.5 \\
\gsm{} & 37.9  & 38.1  & 35.9  & 42    & 38.3  & 38.9  & 38.7  & 39.2  & 39    & 37.5 \\
\drop{} & 37    & 47    & 46.3  & 46.5  & 52.3  & 53.8  & 53.2  & 53.6  & 53.6  & 54.6 \\
\boolq{} & 86.5  & 87    & 86    & 87.7  & 86.5  & 86.9  & 87.4  & 87.2  & 87    & 88 \\
\cola{} & 80.2  & 79.4  & 81.6  & 80.8  & 80.1  & 80.5  & 80.4  & 80.7  & 83.7  & 83.1 \\
\textcolor[rgb]{ .506,  .086,  .055}{\tweet{} (emotion)} & 71.7  & 72.5  & 74.1  & 75.7  & 71.9  & 77.3  & 75.1  & 76.5  & 76.7  & 78.1 \\
\textcolor[rgb]{ .506,  .086,  .055}{\tweet{} (offensive)} & 68.2  & 70.5  & 71.7  & 68.3  & 74.7  & 73    & 72.2  & 73.1  & 75    & 76.3 \\
\midrule
AVG (Held-out) & 45.33 & 58    & 58.84 & 65.01 & 66.44 & 66.59 & 66.46 & 66.57 & 71.13 & 72.93 \\
AVG (All) & 58.79 & 65.1  & 65.54 & 70.06 & 71.85 & 71.96 & 71.82 & 71.68 & 74.9  & 75.68 \\
\bottomrule
\end{tabular}%

%% file: tables/tabulars/full_results/babbage.tex
\begin{tabular}{lcccccc}
\toprule
\multirow{2}[2]{*}{\textbf{Dataset}} & \multirow{2}[2]{*}{\textbf{\rsc{}}} & \multirow{2}[2]{*}{\textbf{\bmsc{}}} & \multirow{2}[2]{*}{\textbf{\cossc{}}} & \multirow{2}[2]{*}{\textbf{\bsrsc{}}} & \textbf{\gistmulti{}} & \textbf{\gistft{}} \\
     &     &     &      &      & \boldmath{}\textbf{$l = 1$}\unboldmath{} & \boldmath{}\textbf{$l = 1$}\unboldmath{} \\
\midrule
\textcolor[rgb]{ .506,  .086,  .055}{\smcalflow{} (CG)} & 0     & 0.3   & 0.6   & 0.8   & 0.9   & 1.2 \\
\textcolor[rgb]{ .506,  .086,  .055}{\smcalflow{} (IID)} & 2.9   & 15.4  & 17.2  & 25.2  & 17.7  & 34.3 \\
\textcolor[rgb]{ .506,  .086,  .055}{\mtop{}} & 2.3   & 45.9  & 46.2  & 52.3  & 50.6  & 58.6 \\
\textcolor[rgb]{ .506,  .086,  .055}{\cogs{} (CG)} & 2.1   & 10.3  & 13.2  & 29.6  & 29.9  & 31.4 \\
\textcolor[rgb]{ .506,  .086,  .055}{\cogs{} (IID)} & 2.2   & 13.7  & 17.4  & 35.4  & 31.1  & 32.7 \\
\qnli{} & 51.7  & 55.6  & 56.8  & 83    & 86.5  & 91.2 \\
\mnli{} & 35.4  & 43.5  & 46.8  & 83.2  & 80.4  & 85.1 \\
\rte{} & 59.9  & 56.3  & 57.4  & 74    & 84.1  & 83 \\
\textcolor[rgb]{ .506,  .086,  .055}{\wanli{}} & 38.7  & 45    & 47.9  & 62.6  & 60.2  & 68.4 \\
\textcolor[rgb]{ .506,  .086,  .055}{\xnli{} (de)} & 34    & 36.5  & 36.8  & 51.6  & 65.9  & 68.4 \\
\textcolor[rgb]{ .506,  .086,  .055}{\xnli{} (ru)} & 32.9  & 38.6  & 39.9  & 39.9  & 52.4  & 55.4 \\
\textcolor[rgb]{ .506,  .086,  .055}{\mednli{}} & 36.7  & 53.4  & 59.3  & 74.3  & 72.4  & 83.2 \\
\ssttwo{} & 90.7  & 88.2  & 87.6  & 94.8  & 92.1  & 94.6 \\
\sstfive{} & 31.4  & 36.8  & 38.7  & 44.4  & 48.6  & 49.4 \\
\textcolor[rgb]{ .506,  .086,  .055}{\rotten{}} & 76.8  & 84.6  & 87.2  & 91    & 90.8  & 90.5 \\
\mrpc{} & 68.4  & 68.9  & 65.9  & 75    & 85    & 87.7 \\
\qqp{} & 56.6  & 56.4  & 64.9  & 83.8  & 82.8  & 87.3 \\
\paws{} & 44.5  & 48.8  & 50.4  & 68.6  & 89.8  & 93.4 \\
\textcolor[rgb]{ .506,  .086,  .055}{\pawsx{} (es)} & 51.9  & 47.2  & 45.7  & 66.5  & 79.3  & 88.7 \\
\textcolor[rgb]{ .506,  .086,  .055}{\pawsx{} (fr)} & 50.6  & 50.6  & 50.9  & 65.9  & 83.3  & 91 \\
\cmsqa{} & 20.9  & 20    & 19.6  & 20.4  & 55.5  & 63.4 \\
\agnews{} & 85.7  & 92.3  & 92.5  & 93.4  & 92.9  & 93.3 \\
\gsm{} & 2.7   & 4.1   & 3.6   & 5     & 2.8   & 4.6 \\
\drop{} & 10.9  & 14.5  & 15.1  & 14    & 24.3  & 30.1 \\
\boolq{} & 64.3  & 68    & 67.8  & 70.3  & 82.8  & 82.8 \\
\cola{} & 68.6  & 64.3  & 67    & 69    & 76.6  & 79.3 \\
\textcolor[rgb]{ .506,  .086,  .055}{\tweet{} (emotion)} & 42.5  & 48.1  & 58.6  & 64.7  & 73.5  & 79.1 \\
\textcolor[rgb]{ .506,  .086,  .055}{\tweet{} (offensive)} & 52.5  & 64.7  & 70.4  & 65.8  & 78    & 76.1 \\
\midrule
AVG (Held-out) & 30.44 & 39.59 & 42.24 & 51.83 & 56.14 & 61.36 \\
AVG (All) & 39.92 & 45.43 & 47.34 & 57.3  & 63.22 & 67.29 \\
\bottomrule
\end{tabular}%

%% file: tables/tabulars/full_results/davinci.tex
\begin{tabular}{lcccccc}
\toprule
\multirow{2}[2]{*}{\textbf{Dataset}} & \multirow{2}[2]{*}{\textbf{\rsc{}}} & \multirow{2}[2]{*}{\textbf{\bmsc{}}} & \multirow{2}[2]{*}{\textbf{\cossc{}}} & \multirow{2}[2]{*}{\textbf{\bsrsc{}}} & \textbf{\gistmulti{}} & \textbf{\gistft{}} \\
     &     &     &      &      & \boldmath{}\textbf{$l = 1$}\unboldmath{} & \boldmath{}\textbf{$l = 1$}\unboldmath{} \\
\midrule
\textcolor[rgb]{ .506,  .086,  .055}{\smcalflow{} (CG)} & 0     & 0.8   & 1.6   & 2.4   & 3.2   & 1.2 \\
\textcolor[rgb]{ .506,  .086,  .055}{\smcalflow{} (IID)} & 0.8   & 12.4  & 17.2  & 29.6  & 22    & 22.8 \\
\textcolor[rgb]{ .506,  .086,  .055}{\mtop{}} & 2.4   & 55.6  & 52.4  & 56.8  & 59.2  & 61.2 \\
\textcolor[rgb]{ .506,  .086,  .055}{\cogs{} (CG)} & 10    & 21.2  & 21.6  & 46.4  & 44.8  & 48.4 \\
\textcolor[rgb]{ .506,  .086,  .055}{\cogs{} (IID)} & 6.4   & 22.4  & 24.8  & 44    & 44.4  & 40 \\
\qnli{} & 45.2  & 57.2  & 52    & 82    & 84.4  & 92.4 \\
\mnli{} & 55.6  & 62.8  & 60    & 84.8  & 82.4  & 83.2 \\
\rte{} & 77.2  & 71.6  & 71.6  & 80    & 88.4  & 85.2 \\
\textcolor[rgb]{ .506,  .086,  .055}{\wanli{}} & 49.2  & 50.8  & 52.8  & 65.2  & 62.4  & 71.6 \\
\textcolor[rgb]{ .506,  .086,  .055}{\xnli{} (de)} & 42.8  & 46.8  & 44.4  & 52.4  & 73.2  & 69.6 \\
\textcolor[rgb]{ .506,  .086,  .055}{\xnli{} (ru)} & 41.6  & 45.6  & 43.6  & 43.2  & 59.6  & 60.8 \\
\textcolor[rgb]{ .506,  .086,  .055}{\mednli{}} & 61.6  & 75.6  & 72.8  & 83.2  & 78.4  & 84 \\
\ssttwo{} & 94.8  & 88.4  & 89.2  & 95.6  & 94    & 94 \\
\sstfive{} & 45.2  & 50.8  & 52    & 47.6  & 51.2  & 54.8 \\
\textcolor[rgb]{ .506,  .086,  .055}{\rotten{}} & 93.2  & 91.2  & 94.8  & 94    & 94    & 94 \\
\mrpc{} & 71.6  & 68.8  & 62.4  & 78    & 85.2  & 89.2 \\
\qqp{} & 70.4  & 76.8  & 78.8  & 85.6  & 83.2  & 86 \\
\paws{} & 67.6  & 55.6  & 60    & 80.8  & 90.4  & 94.4 \\
\textcolor[rgb]{ .506,  .086,  .055}{\pawsx{} (es)} & 64.4  & 59.2  & 55.2  & 70.4  & 79.6  & 84.4 \\
\textcolor[rgb]{ .506,  .086,  .055}{\pawsx{} (fr)} & 65.6  & 59.6  & 65.6  & 67.6  & 82.8  & 88.8 \\
\cmsqa{} & 72.8  & 65.6  & 67.2  & 66.8  & 77.6  & 75.2 \\
\agnews{} & 86    & 94.8  & 93.6  & 92    & 93.6  & 92.8 \\
\gsm{} & 32.8  & 30    & 33.6  & 37.2  & 36.8  & 35.2 \\
\drop{} & 36    & 38    & 42.8  & 37.6  & 49.6  & 49.6 \\
\boolq{} & 82.8  & 84    & 88    & 88    & 91.6  & 88 \\
\cola{} & 73.2  & 74.8  & 78.8  & 77.6  & 77.2  & 75.6 \\
\textcolor[rgb]{ .506,  .086,  .055}{\tweet{} (emotion)} & 58    & 62.8  & 69.2  & 64.8  & 66.8  & 79.6 \\
\textcolor[rgb]{ .506,  .086,  .055}{\tweet{} (offensive)} & 68.8  & 69.2  & 70.4  & 71.6  & 78.5  & 78.1 \\
\midrule
AVG (Held-out) & 40.34 & 48.09 & 49.03 & 56.54 & 60.64 & 63.18 \\
AVG (All) & 52.71 & 56.87 & 57.73 & 65.19 & 69.09 & 70.72 \\
\bottomrule
\end{tabular}%